\documentclass[10pt,twocolumn,letterpaper]{article}

\usepackage[pagenumbers]{cvpr} 

\usepackage{graphicx}
\usepackage{amsmath}
\usepackage{mathtools}
\usepackage{booktabs}
\usepackage{comment}
\usepackage{xcolor}
\usepackage{float}
\usepackage{subcaption}
\usepackage{pifont}\usepackage{multirow}
\usepackage{bigdelim}
\usepackage{stfloats}

\definecolor{cvprblue}{rgb}{0.21,0.49,0.74}
\usepackage[pagebackref,breaklinks,colorlinks,citecolor=cvprblue]{hyperref}

\def\paperID{16415} \def\confName{CVPR}
\def\confYear{2025}

\DeclareMathOperator*{\argmin}{arg\,min}

\newcommand{\cmark}{\ding{51}}\newcommand{\xmark}{\ding{55}}

\definecolor{ForestGreen}{HTML}{009B55}

\newcommand{\OURS}{GaussianSpeech}

\title{\OURS: Audio-Driven Gaussian Avatars}

\author{Shivangi Aneja\textsuperscript{\rm 1}
$\quad$
Artem Sevastopolsky\textsuperscript{\rm 1}
$\quad$
Tobias Kirschstein\textsuperscript{\rm 1}
$\quad$
Justus Thies\textsuperscript{\rm 2,}\textsuperscript{\rm 3} \vspace{0.3em}\\
$\quad$
Angela Dai\textsuperscript{\rm 1}
$\quad$
Matthias Nie\ss ner\textsuperscript{\rm 1} \vspace{0.3em}\\
{\normalsize \textsuperscript{\rm 1}Technical University of Munich} \quad
{\normalsize \textsuperscript{\rm 2}MPI-IS, T{\"u}bingen} \quad
{\normalsize \textsuperscript{\rm 3}TU Darmstadt} \quad
}

\begin{document}

\twocolumn[{\renewcommand\twocolumn[1][]{#1}\vspace{-0.5cm}
\maketitle
\begin{center}
    \centering
    \vspace{-0.7cm}
    \includegraphics[width=0.95\linewidth]{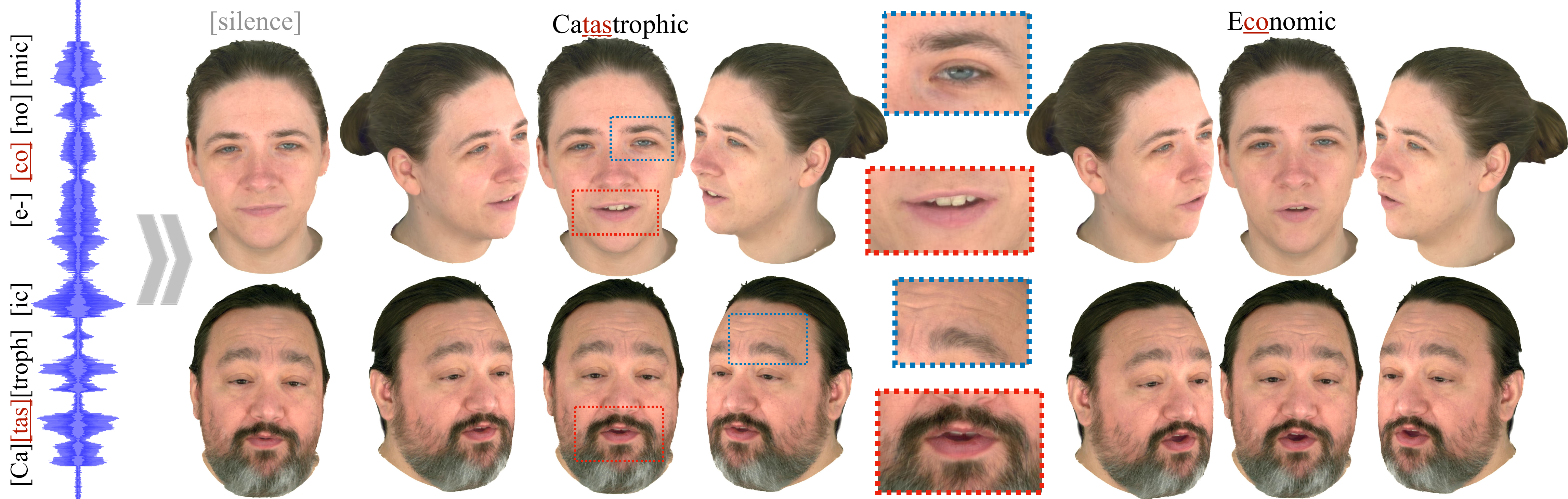}
    \vspace{-0.2cm}
    \captionof{figure}{Given input speech signal, \OURS{} can synthesize photorealistic 3D-consistent talking human head avatars. Our method can generate realistic and high-quality animations, including mouth interiors such as teeth, wrinkles, and specularities in the eyes. We handle diverse facial geometry, including hair buns and mustaches/beards, while effectively synchronizing to the audio signal.
    }
    \label{fig:teaser}
\end{center}}]

\begin{abstract}
\vspace{-0.25cm}

We introduce \OURS{}\footnote{Project Page: \url{https://shivangi-aneja.github.io/projects/gaussianspeech}}, a novel approach that synthesizes high-fidelity animation sequences of photo-realistic, personalized 3D human head avatars from spoken audio.
To capture the expressive, detailed nature of human heads, including skin furrowing and finer-scale facial movements, we propose to couple speech signal with 3D Gaussian splatting to create realistic, temporally coherent motion sequences.
We propose a compact and efficient 3DGS-based avatar representation that generates expression-dependent color and leverages wrinkle- and perceptually-based losses to synthesize facial details, including wrinkles that occur with different expressions.
To enable sequence modeling of 3D Gaussian splats with audio, we devise an audio-conditioned transformer model capable of extracting lip and expression features directly from audio input.
Due to the absence of high-quality dataset of talking humans in correspondence with audio, we captured a new large-scale multi-view dataset of audio-visual sequences of talking humans with native English accents and diverse facial geometry.
\OURS{} consistently achieves state-of-the-art quality with visually natural motion, while encompassing diverse facial expressions and styles.
 
\end{abstract}

\vspace{-0.2cm} 

\section{Introduction}
\label{sec:intro}

Generating animated sequences of photorealistic 3D head avatars from spoken audio is important for many graphics applications, including immersive telepresence, movies, and virtual assistants.
In particular, rendering photorealistic views of such animated avatars from various viewpoints is crucial for realistic, immersive digital media, for instance, telepresence to a meeting room requires a photorealistic appearance for all viewpoints of the people in the room, or AR/VR where users can freely change their viewpoint.

Creating such photorealistic animated 3D avatars from audio remains challenging, as it requires maintaining photorealistic fidelity throughout the animation sequence, as well as from various viewpoints.
Existing work thus focuses on addressing these objectives independently; various works focus on re-enacting videos in the 2D domain \cite{guo2021adnerf, zhang2024learning, shen2022dfrf, yao2022dfa,Bai_2023_CVPR, ye2023geneface, li2023ernerf, sd_nerf, peng2023synctalk, GaussianTalker2024, talkingGaussian2024}, creating front-view video animations, while others focus on animating 3D face geometry from audio~\cite{richard2021meshtalk,faceformer2022, xing2023codetalker, Thambiraja_2023_ICCV}.
In contrast, we aim to create innately 3D audio-driven avatars enabling 3D-consistent, free-viewpoint photorealistic synthesis needed for immersive digital communication.

In order to characterize audio-driven 3D animation of a person from multi-view input, we propose to represent animated head sequences with explicit 3D Gaussian points, leveraging the detailed and expressive representation space of 3D Gaussian Splatting (3DGS)~\cite{kerbl3Dgaussians}.
3DGS offers a flexible representation capable of handling complex and irregular facial geometry and appearance (e.g., different skin tones, beard, skin creasing) and real-time rendering, making it a well-suited choice for facial animation.

Thus, we design an efficient, personalized 3D Gaussian avatar representation from multi-view input observations of a person, containing relatively few Gaussian splats in order to make sequence modeling of photorealistic 3DGS tractable and allowing us to operate at real-time rendering rates.
This is achieved through learning expression- and view-dependent color, and our losses focusing on perceptual face quality using a face recognition network, as well as focusing on fine-scale details through wrinkle detection.
 
Our efficient, high-quality avatar can handle the nuances of the facial geometry, like skin tone variation and dynamic wrinkles. 
We then use this person-specific avatar to guide audio-driven head animation, enabled by our transformer-based sequence model.
We learn lip motion features and wrinkle features directly from audio to obtain expression input to train our transformer model, enabling photorealistic generation of a coherent animation sequence.

To create high-fidelity, audio-driven animated 3D head avatars, we require high-resolution multi-view data paired with high-quality audio recordings.
Existing multiview datasets~\cite{kaisiyuan2020mead,kirschstein2023nersemble} unfortunately lack either high-quality video or high-quality audio captures.  In the absence of large-scale and high-quality paired audio-multiview data of people speaking, we collected a new multiview dataset with 16 cameras for 6 native English participants captured at 30 fps and 3208x2200 resolution with overall recordings of $\sim$3.5 hours, an order of magnitude larger than the existing datasets. We will make the dataset and the corresponding 3D face trackings publicly available for research purposes.

\medskip
In this work, we introduce a novel approach for high-fidelity and multi-view consistent sequence animation of photorealistic 3D head avatars for content creation applications. To summarize, our contributions are:

\begin{itemize}    
    \item The first transformer-based sequence model for audio-driven head animation synthesis of a lightweight 3DGS based avatar. By animating our optimized 3DGS avatar directly with our transformer model, we achieve temporally coherent animation sequences while characterizing fine-scale face details and speaker-specific style.
    
    \item A new high-quality audio-video dataset, comprising high-resolution 16-view dataset of 6 native English speakers (Standard American \& British). The dataset has  a total of 2500 sequences, with overall recordings of $\sim$3.5 hours.
\end{itemize} 
\section{Related Work}
\label{sec:related_work}

Audio-driven facial animation plays an important role in digital media. Here we discuss audio-driven animation methods  generating different output representations.

 \subsection{2D-Based Methods.}  
There is a large corpus of works in the field of 2D audio-driven facial animation operating on monocular RGB videos, synthesizing 2D sequences directly~\cite{tian2024emo, Mukhopadhyay_2024_WACV, wang2021one, xu2024vasa1, lipformer, guan2023stylesync, VideoReTalking, chung2017said, Wiles18, chen2018lip, suwajanakorn2017obama, wave2lip2020, chen2019hierarchical, lipgan2019, make_it_talk2020, vougioukas2018endtoend, vougioukas2019realistic, zhou2018talking, cascade2020, chen2020talking, emotion22, gururani2022space, shen2023difftalk, stypulkowski2022diffused, Gupta_2023_WACV, xie2024x}. However, these methods operate in pixel space and can produce very limited side views. Another line of work also operating on frontal RGB videos but using intermediate 3D representations are based on 3DMMs~\cite{zhang2021flow, song2020everybodys, thies2020nvp, ji2021audio-driven, Doukas_2021_ICCV, tang2022memories}. Although these methods generate photorealistic results, they use 3DMMs as a proxy to improve the animation quality and are still limited to frontal and limited side views. In contrast, we model head avatars with explicit 3D Gaussian points, thus, enabling simultaneous free-viewpoint rendering for different viewpoints which is critical for telepresence applications.

 \subsection{Parametric Model Based Methods.}
Another promising line of work is to animate 3D facial geometry directly. A vast majority of these works model speech-conditioned animation for either artist-designed template meshes~\cite{karras2017audio, VOCA2019, richard2021meshtalk, faceformer2022,  xing2023codetalker, Thambiraja_2023_ICCV, EMOTE, thambiraja20233diface} or blendshapes for 3D parametric head model~\cite{Peng_2023_ICCV, aneja2023facetalk}. While these methods can faithfully match facial motion with the speech signal and can be rendered from different viewpoints, they do not model any appearance or texture information and cannot handle complex and irregular facial geometry. The synthesized animations, therefore, do not look realistic. Compared to these, our method optimizes a 3DGS-based avatar and models appearance using expression and view-dependent color, generating photorealistic results.

\subsection{Radiance Fields Based Methods.}
Recent speech-driven animation methods based on radiance fields \cite{guo2021adnerf, yao2022dfa, shen2022dfrf, liu2022semantic, ye2023geneface, li2023ernerf, peng2023synctalk} have gained popularity due to their ability to model directly from images. Neural Radiance Fields (NeRF)~\cite{mildenhall2020nerf} possess the capability to render a scene from arbitrary viewpoints, however, existing audio-driven methods utilizing NeRF are designed for monocular videos.

Concurrent to ours, few recent works~\cite{GaussianTalker2024, talkingGaussian2024, he2024emotalk3d} leverage 3DGS~\cite{kerbl3Dgaussians} for generating audio-driven talking heads. GaussianTalker~\cite{GaussianTalker2024} and TalkingGaussian~\cite{talkingGaussian2024} focus on improving the rendering speed for monocular videos. EmoTalk3D~\cite{he2024emotalk3d} can synthesize multi-view renders, however these methods generate sequences frame-by-frame, thus suffer from jitter and scaling artefacts. In contrast, our method synthesizes multi-view consistent and temporally smooth results, including fine-scale details like dynamic wrinkles, by leveraging a transformer-based sequence model and an efficient 3DGS-based avatar.

 \section{Multi-View Audio-Visual Dataset}

\begin{figure}[t!]
 \vspace{-0.3cm}
 \centering
 \includegraphics[width=1.0\linewidth]{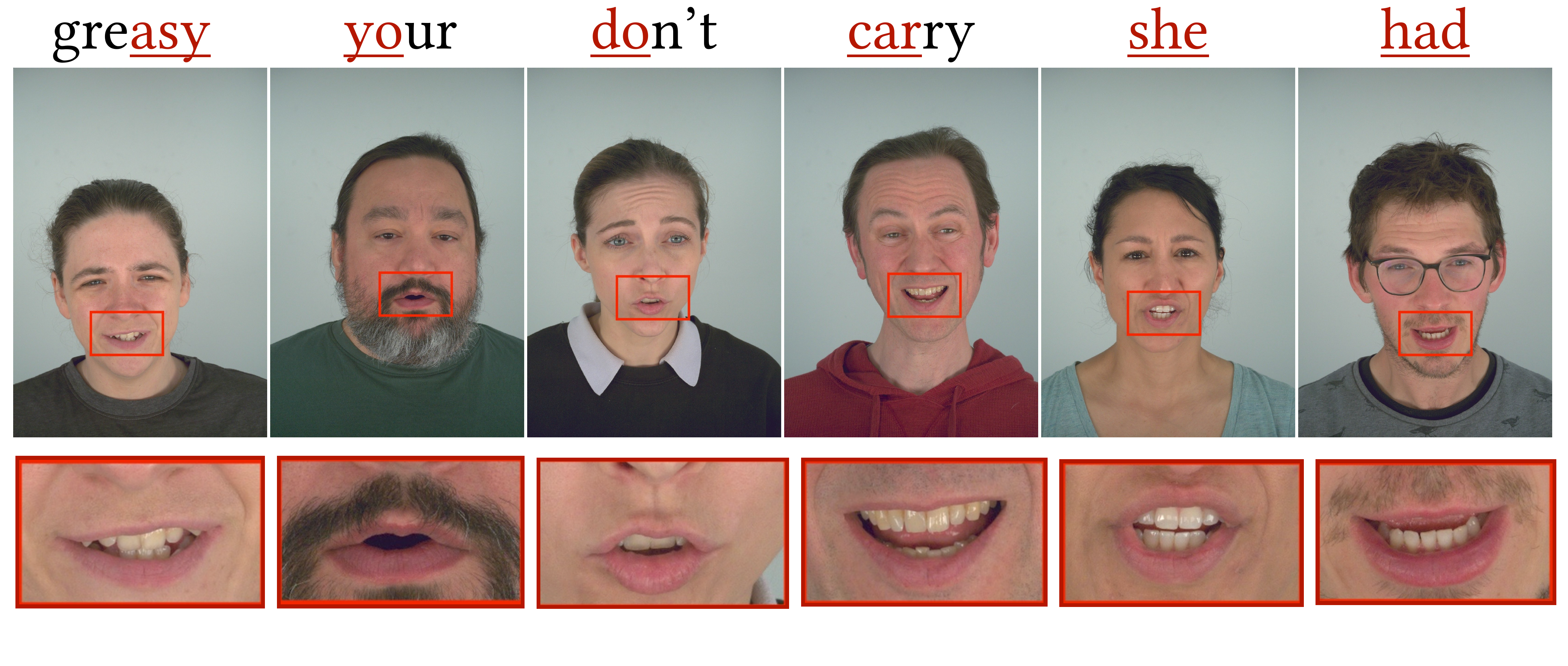}
 \vspace{-0.8cm}
 \caption{Random frames selected for each participant (top) from the dataset and corresponding zoom-in for the mouth region (bottom). We captured a gender-balanced dataset of native speakers with different English accents and diverse facial geometry including different skin tones, beard and glasses to maximize diversity.}
 \label{fig:dataset_particpants}
\end{figure}

We collected a novel dataset consisting of six native English speakers captured using a multiview rig of 16 cameras (see Supp.). We record sequences at 30 FPS at 3208 x 2200 resolution. To achieve quality and diversity, we specifically capture native English speakers with different accents, including American, British, and Canadian. We selected participants aged 20-50 with different genders and facial geometry including beard and glasses to increase the diversity, see Fig.~\ref{fig:dataset_particpants}. 
We collected 415 sequences for every subject, leading to an overall recording time of 30-35 minutes for each of the 16 cameras. The spoken sentences are chosen from the TIMIT~\cite{timit} corpus to maximize the phonetic diversity. Our dataset stands out from the existing datasets in terms of quality and quantity.

While certain datasets with audio-visual talking faces exist, they are limited in quality. The RAVDESS dataset~\cite{livingstone_2018_1188976} contains a set of native speakers, but it has only 2 unique sequences per participant with North American accent, while we captured three different English accents and 415 unique sentences. The MEAD dataset~\cite{kaisiyuan2020mead} captured the participants with 250 unique sentences per participant. However, they focus on emotional speech synthesis due to which they capture only 40 unique natural expression/emotion per participant at a relatively lower resolution. The Nersemble~\cite{kirschstein2023nersemble} dataset captures the participants at high resolution, but it only contains 10 audio sequences per participant. Closest to ours is MultiFace~\cite{wuu2022multiface}, which captured participants in a spherical rig of 150 cameras; however, it captured only 50 audio sequences per participant. Our dataset contains 415 sequences for every subject at high resolution, an order of magnitude larger than existing datasets, see Tab.~\ref{tab:dataset_details}. We plan to release our entire dataset to the research community.

\begin{table}[bh!]
    \begin{center}
    \resizebox{1.0\linewidth}{!}{
    \begin{tabular}{c c c c c c}
        \toprule
        Dataset & {\# Cam} &  {\# Unique} & {Resolution} & {Duration} & {Native}  \\
        &  &  {Sentences} &  & (in minutes/camera) & \\ 
        \toprule
        RAVDESS~\cite{livingstone_2018_1188976} & 1 & 2 & 1920 x 1080 & 0.1 min  & {\textcolor{ForestGreen}{\cmark}} \\
        MEAD~\cite{kaisiyuan2020mead} & 8 & 250  & 1920 x 1080 & 20 min &  {\textcolor{brown}{\xmark}} \\
        EmoTalk3D~\cite{he2024emotalk3d} & 11 & N/A  & 512 x 512 & 20 min &  {\textcolor{brown}{\xmark}} \\
        Nersemble~\cite{kirschstein2023nersemble} & 16 & 10 & \textbf{3208} x \textbf{2200} & 1 min & {\textcolor{brown}{\xmark}} \\
        MultiFace~\cite{wuu2022multiface} & \textbf{150} & 50 & 2048 x 1334 & 4 min & {\textcolor{brown}{\xmark}} \\
        \hline
        Ours & 16 & \textbf{415} & \textbf{3208} x \textbf{2200} & \textbf{35} \textbf{min} &  {\textcolor{ForestGreen}{\cmark}} \\
        \bottomrule
    \end{tabular}}
    \end{center}
    \vspace{-0.6cm}
    \caption{Existing Audio-Video Dataset Comparison per participant in the datasets. Compared to existing datasets, ours is an order of magnitude larger and higher resolution. All datasets are captured at standard 30 fps.}
    \label{tab:dataset_details}
    \vspace{-0.5cm}    
\end{table}

 \begin{figure}[t]
     \centering
     \includegraphics[width=1.0\linewidth]{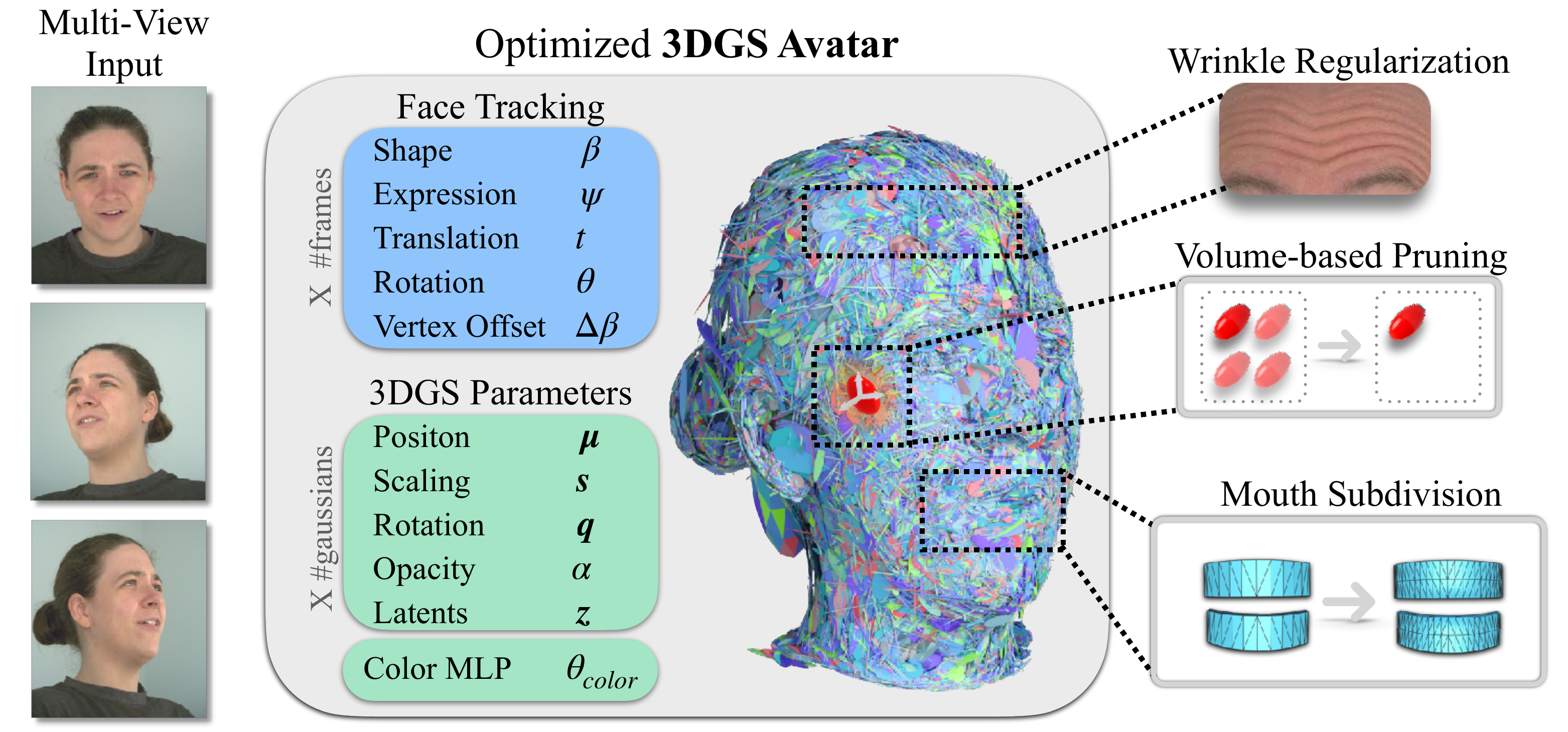}
     \vspace{-0.6cm}
     \caption{Person-specific 3D Avatar: We compute 3D face tracking and bind 3D Gaussians to the triangles of the tracked FLAME mesh. We apply volume-based pruning to prevent optimization to generate large amount of Gaussians, and apply subdivision of mesh triangles in the mouth region. We train color MLP $\theta_\textrm{color}$ to synthesize expression \& view dependent color. We apply wrinkle regularization and perceptual losses to improve photorealism.}
     \label{fig:avatar_init_gsavatar}
 \end{figure}

\section{Method}
Our method operates in two stages. First, we develop a lightweight and high-quality avatar initialization based on GaussianAvatars (Sec.~\ref{method:avatara_init}). Next, we train a transformer-based sequence model to animate our initialized avatar conditioned on personalized audio features (Sec.~\ref{method:avatar_animation}). Since our method requires 3D face tracking, we compute them from our multiview sequence dataset, similar to~\cite{qian2023gaussianavatars}. 

\subsection{Avatar Initialization}\label{method:avatara_init}
We propose an efficient optimization strategy to compute a 3DGS-based Gaussian avatar representation.
We found that naively training GaussianAvatar~\cite{qian2023gaussianavatars} generates blurred/low-quality textures, especially, for scenarios with rapid facial movement like faster talking speed/head motion.
In addition, GaussianAvatar can not effectively handle dynamic wrinkles.
Therefore, we introduce expression-dependent colors and propose several regularizations to improve quality of our avatars described below and shown in Fig.~\ref{fig:avatar_init_gsavatar}.

{\textbf{Volume-Based Pruning.}} We modify the pruning strategy used by GaussianAvatar. Instead of pruning 3D Gaussian splats based on a given opacity threshold $\boldsymbol{\epsilon}_\textrm{opacity}$, we select top 25,000 Gaussians with maximum opacity and 3D Gaussian's scale volume combined at every pruning step as
\begin{equation}
    \mathcal{G}_{i} = \sigma_{i} \cdot (s_x \cdot s_y \cdot s_z),
\end{equation}
where $\sigma_{i}$ refers to $i^{th}$ Gaussian's opacity and $s_x, s_y, s_z$ refers to its scale along x, y, and z axis. Even when the optimization generates excessive splats during densification, this top-k pruning ensures that the optimized avatar does not contain too many 3D Gaussian splats. However, this leads to degradation in quality by removing small transparent 3D splats and generates blurry results. We, thus, propose to add additional regularizations to improve quality.

{\textbf{Expression-dependent Color.}} 
Instead of learning SH Color for 3D Gaussians, our method generates color with a lightweight two-layer color MLP $\boldsymbol{\theta}_\textrm{color}$ to faithfully synthesize dynamic wrinkles.
Given a FLAME~\cite{flame_siggraphAsia2017} expression code $\boldsymbol{\psi}$ and viewing direction $\boldsymbol{v}$, we synthesize view- and expression-dependent color $\boldsymbol{c}_{i}$ as: 
\begin{equation}
    \boldsymbol{c}_{i} = \boldsymbol{\theta}_\textrm{color} (\boldsymbol{\psi}; \boldsymbol{z}_i; \boldsymbol{v} ).
\end{equation}
Note that we additionally learn per Gaussian latent features $\boldsymbol{z}_i$ for sharper colors.

{\textbf{Perceptual Losses.}} To improve the sharpness of the color generated by $\boldsymbol{\theta}_\textrm{color}$, we add a global and patch-based perceptual loss. The global perceptual loss $\mathcal{L}_\textrm{global}$ is based on the content and style features of the pre-trained face recognition model ArcFace~\cite{arcface}.
The content loss $\mathcal{L}_\textrm{content}$ and style loss $\mathcal{L}_\textrm{style}$ are defined as:
\vspace{-0.1cm}
\begin{align}
    \mathcal{L}_\textrm{content} &= \sum_{k=1}^{K} \big|\big| \phi_{k}(I_\textrm{render}) - \phi_{k}(I_{gt}) \big|\big|_1 , \\
    \mathcal{L}_\textrm{style} &= \sum_{k=1}^{K} \big|\big| \mathcal{G}_{k}(I_\textrm{render}) - \mathcal{G}_{k}(I_{gt}) \big|\big|_1 ,
\end{align}
\vspace{-0.2cm}

\noindent where $\phi_{k}$ and $\mathcal{G}_{k}$ refer to the feature maps and Gram matrices~\cite{johnson2016perceptual} for the layer $k$ respectively.  $I_\textrm{render}$ and $I_\textrm{gt}$ refer to the rendered and ground-truth multiview image. 

\begin{equation}
 \mathcal{L}_\textrm{global} = \mathcal{L}_\textrm{content} + \mathcal{L}_\textrm{style}.
 \label{eq:perceptual}
\end{equation}

$\mathcal{L}_\textrm{global}$ explained above improves the quality of the texture globally, however, 
it shows limited improvements for fine-scale skin areas and less observed regions like the mouth interior. We, therefore, employ a VGG-based loss on local image patches based on content features of the pre-trained VGG backbone as:
\begin{equation}
    \mathcal{L}_\textrm{patch} = \frac{1}{J}  \sum_{j=1}^{J} \sum_{k=1}^{K}   \big|\big| \zeta_{k}(I_\textrm{render}^{j}) - \zeta_{k}(I_\textrm{gt}^{j}) \big|\big|_1,
\end{equation}
where   $I_\textrm{render}^{j}$ and $I_\textrm{gt}^{j}$ refer to the $j^{th}$ local patch regions from  the rendered and ground-truth multiview images. We use $128 \times 128$ patches  and sample 16 local patches uniformly for the facial area by employing alpha matting.

{\textbf{Wrinkle Regularization.}} 
Naive optimization of GaussianAvatar~\cite{qian2023gaussianavatars} cannot represent skin creasing and fine-scale wrinkles, since it learns a constant color for the avatar, irrespective of facial expression.
To overcome this, we introduce a lightweight color MLP $\theta_{color}$ that can generate expression-dependent wrinkles.
We employ a novel wrinkle feature loss $\mathcal{L}_\textrm{wrinkle}$ which focuses on refining dynamic wrinkles.
Specifically, we run an off-the-shelf wrinkle detector~\cite{wrinkle_detection} to extract wrinkle features and apply a content loss on its feature detection backbone during optimization:
\vspace{-0.2cm}
\begin{equation}
    \mathcal{L}_\textrm{wrinkle} = \sum_{k=1}^{K} \big|\big| \Psi_{k}(I_\textrm{render}) - \Psi_{k}(I_\textrm{gt}) \big|\big|_1.
    \label{eq:wrinkle}
    \vspace{-0.1cm}
\end{equation}
Note that our method synthesizes wrinkles faithfully for avatars whose captured data includes dynamic wrinkles when speaking; if the avatar did not display wrinkles during speech, our method will not generate them.

{\textbf{Mouth Region Subdivision.}} Since the mouth interior (especially teeth) is less frequently observed compared to other facial regions, the standard 3DGS-based densification cannot generate sufficient Gaussians for the mouth to synthesize high quality results. To address this, before optimization, we subdivide the triangles which are used to initialize the Gaussians corresponding to the teeth in the FLAME mesh using a uniform four-way subdivision. By doing so, we begin with a high density of Gaussians for the teeth, compensating for low gradient magnitude in this area, ensuring that teeth appear detailed and realistic.

To summarize, we optimize our 3DGS-based avatar using $\mathcal{L}_\textrm{total}$ loss as:
\vspace{-0.1cm}
\begin{equation}
\begin{split}
    \mathcal{L}_\textrm{total} = \mathcal{L}_\textrm{rgb} + \lambda_\textrm{pos} \mathcal{L}_\textrm{position} +  \lambda_\textrm{s} \mathcal{L}_\textrm{scaling} \\
    + \, \lambda_\textrm{g} \mathcal{L}_\textrm{global} + \, \lambda_\textrm{p} \mathcal{L}_\textrm{patch} +  \lambda_\textrm{w} \mathcal{L}_\textrm{wrinkle},
\end{split}
\label{eq:avatar_equation}
\end{equation}

\noindent where $\mathcal{L}_\textrm{rgb}, \mathcal{L}_\textrm{position}, \mathcal{L}_\textrm{scaling}$ are defined in~\cite{qian2023gaussianavatars} (also explained in Supp. doc).

\begin{figure*}[tp]
    \centering
    \includegraphics[width=1.0\linewidth]{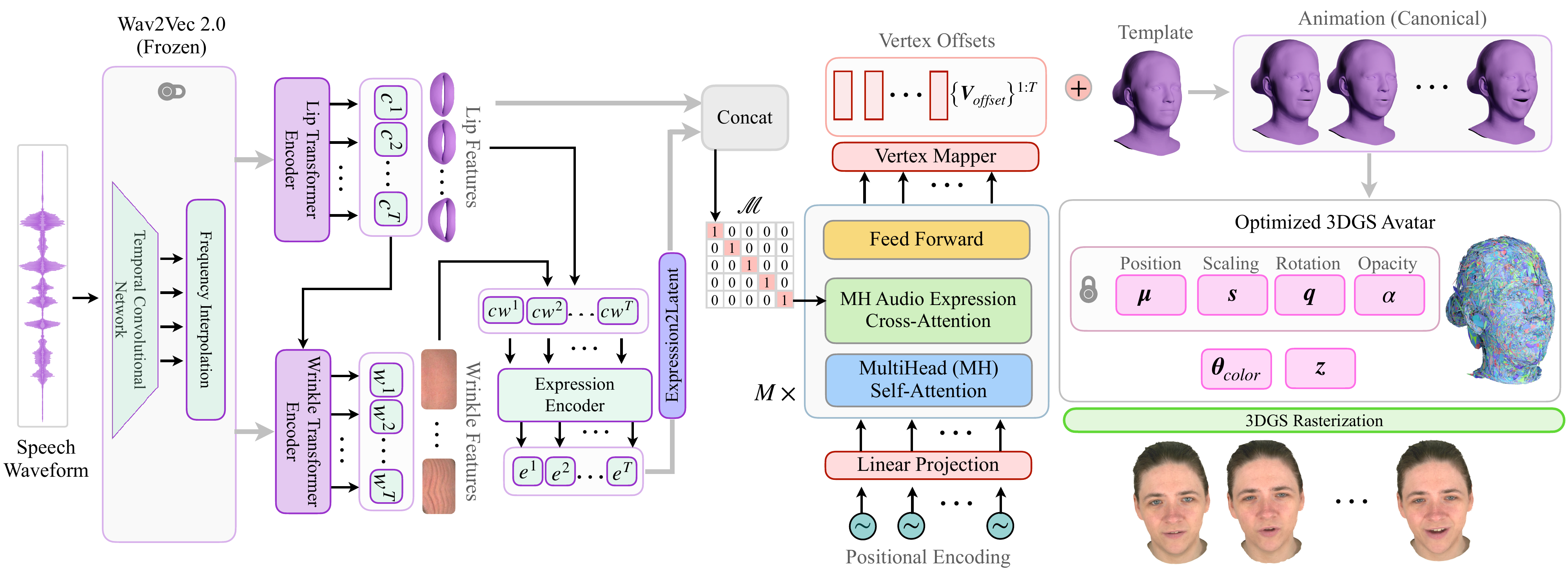}
    \vspace{-0.7cm}
    \caption{Method Overview. From the given speech signal, \OURS{} uses Wav2Vec 2.0~\cite{baevski2020wav2vec} encoder to extract generic audio features and maps them to personalized lip feature embeddings $\boldsymbol{c}^{1:T}$ with Lip Transformer Encoder and wrinkle features $\boldsymbol{w}^{1:T}$ with Wrinkle Transformer Encoder. Next, the Expression Encoder synthesizes FLAME expressions $\boldsymbol{e}^{1:T}$ which are then projected via Expression2Latent MLP and concatenated with $\boldsymbol{c}^{1:T}$ for input to the motion decoder. The motion decoder employs a multi-head transformer decoder~\cite{vaswani2023attention} consisting of Multihead Self-Attention, Cross-Attention, and Feed Forward layers. The concatenated lip-expression features are fused into the decoder via cross-attention layers with alignment mask $\mathcal{M}$. The decoder then predicts FLAME vertex offsets $\{ \boldsymbol{V}_\textrm{offset} \}^{1:T}$ which gets added to the template mesh $\boldsymbol{T}$ to generate vertex animation in canonical space. During training, these are then fed to our optimized 3DGS avatar (Sec.~\ref{method:avatara_init}) and the color MLP $\boldsymbol{\theta}_\textrm{color}$ and gaussian latents $\boldsymbol{z}$ are further refined via re-rendering losses ~\cite{kerbl3Dgaussians}.}
    \label{fig:method_overview}
\end{figure*}

\subsection{Sequence Model Training}\label{method:avatar_animation}
\OURS{} performs high-fidelity and temporally-consistent generative synthesis of avatar motion sequences, conditioned on  audio signal. 
To characterize complex face motions and fine-scale movements like dynamic wrinkles, we employ a transformer-based sequence model. We predict mesh animations with our sequence model and refine the dynamic motion attributes of the 3D Gaussian Splats of our optimized avatar to be consistent with audio features. 
An overview of our approach is illustrated in Fig.~\ref{fig:method_overview}.

\vspace{0.1cm}
\textbf{Audio Encoding.}
We employ the state-of-the-art pre-trained speech model Wav2Vec 2.0~\cite{baevski2020wav2vec} to encode the audio signal. Specifically, we use the audio feature extractor made up of temporal convolution layers (TCN) to extract audio feature vectors $ \{ a_i \}_{i=1}^{N_a}$ from the raw waveform, followed by a \textit{Frequency Interpolation} layer to align the input audio signal $ \{ a_i \}_{i=1}^{N_a}$ (captured at frequency $f_a$ = 16kHz) with our dataset $ \{ a_i \}_{i=1}^{N_e}$ (captures at framerate $f_e$ = 30FPS). 

\vspace{0.1cm}
\textbf{Lip Features.} A stacked multi-layer \textit{Lip Transformer Encoder} processes these resampled audio features and predicts personalized lip content feature vectors $\boldsymbol{c}^{1:T}$. To avoid learning spurious correlation between upper face motion and audio, the Lip Transformer Encoder is trained with only lip vertices from the FLAME mesh with L2-reconstruction loss autoregressively as:
\begin{equation}
    \mathcal{L}_\textrm{lip} = \sum_{n=1}^{N} \Big( \sum_{t=1}^{T} \big|\big| \boldsymbol{l}_\textrm{gt}^{t} - \boldsymbol{l}_\textrm{pred}^{t} \big|\big|_2 \Big)_{n},
\end{equation}
where $T$ refers to the number of frames per sequence and $N$ total sequences, $\boldsymbol{l}_\textrm{gt}$ and $\boldsymbol{l}_\textrm{pred}$ refer to the ground truth and predicted lip vertices, respectively. 

\vspace{0.1cm}
\textbf{Wrinkle Features.} Similarly, our \textit{Wrinkle Transformer Encoder} conditioned on audio and lip features predicts personalized wrinkle feature vectors $\boldsymbol{w}^{1:T}$. The Wrinkle Transformer Encoder is trained with wrinkle features extracted using a wrinkle detector~\cite{wrinkle_detection} from the RGB frames as:
\vspace{-0.2cm}
\begin{equation}
    \mathcal{L}_\textrm{wrinkle} = \sum_{n=1}^{N} \Big( \sum_{t=1}^{T} \big|\big| \boldsymbol{w}_\textrm{gt}^{t} - \boldsymbol{w}_\textrm{pred}^{t} \big|\big|_2 \Big)_{n},
\end{equation}
\vspace{-0.3cm}

\noindent where $\boldsymbol{w}_\textrm{gt}$ and $\boldsymbol{w}_\textrm{pred}$ refer to the ground truth and predicted wrinkle vertices respectively. 

\vspace{0.1cm}
\textbf{Expression Features.}
Using personalized lip features $\boldsymbol{c}^{1:T}$ and wrinkle features $\boldsymbol{w}^{1:T}$ obtained above, we train the \textit{Expression Encoder} $\mathcal{E}_\textrm{exp}$. Specifically, we concatenate lip and wrinkle features to obtain combined features $\boldsymbol{cw}^{1:T} = \left[ \boldsymbol{c}^{1:T}; \boldsymbol{w}^{1:T} \right]$.
These combined features are fed to our Expression Encoder which predicts FLAME expressions as $\boldsymbol{e}^{1:T}_\textrm{pred}$ = $\mathcal{E}_\textrm{exp}(\boldsymbol{cw}^{1:T})$ and is trained with:
\begin{equation}
    \mathcal{L}_\textrm{expr} = \sum_{n=1}^{N} \Big( \sum_{t=1}^{T} \big|\big| \boldsymbol{e}_\textrm{gt}^{t} - \boldsymbol{e}_\textrm{pred}^{t} \big|\big|_2 \Big)_{n},
\end{equation}
where $\boldsymbol{e}_\textrm{gt}$ and $\boldsymbol{e}_\textrm{pred}$ refers to the ground truth and predicted FLAME expression parameters, respectively.

\vspace{0.1cm}
\textbf{Audio-Conditioned Animation.}
We train a transformer decoder~\cite{vaswani2023attention} network to synthesize mesh \textit{Vertex Offsets} $\big\{ \boldsymbol{V}_\textrm{offset} \big\}^{1:T}$, where $T$ refers to the number of frames in a sequence. During training, we first project the predicted expression parameters $\boldsymbol{e}^{1:T}_\textrm{pred}$ via the \textit{Expression2Latent MLP} $\mathcal{E}$ to the latent space of our model and concatenate it with lip features $\boldsymbol{c}^{1:T}$ to obtain combined lip-expression motion features $\boldsymbol{m}^{1:T} = \left[ \boldsymbol{c}^{1:T}; \mathcal{E}(\boldsymbol{e}^{1:T}) \right]$.

These motion features $\boldsymbol{m}^{1:T}$ are then processed through transformer decoder, and the \textit{Vertex Mapper MLP} to synthesize Vertex Offsets $\big\{ \boldsymbol{V}_\textrm{offset} \big\}^{1:T}$ in canonical space. We leverage a look-ahead binary target mask $\mathcal{T} \in  \mathbb{R}^{N \times N}$ in the multi-head self-attention layer to prevent the model from peeking into the future frames.
The $(i,j)^{th}$ element of the matrix with $1 \leq i,j \leq N $ is:
\begin{equation}
      \mathcal{T}_{ij} =   
    \begin{cases}
        True \quad \text{if } i \leq j\\
        False \quad \text{else}\\ 
    \end{cases}
\end{equation}
\vspace{-0.2cm}

Input motion features $\boldsymbol{m}^{1:T}$ are fused into the transformer with the multi-head audio expression cross-attention layer via the alignment mask $\mathcal{M}$. The binary mask $\mathcal{M} \in  \mathbb{R}^{N \times N}$ is a Kronecker delta function $\delta_{ij}$ such that the motion features for $i^{th}$ timestamp attend to vertex features at the $j^{th}$ timestamp if and only if $i=j$:
\begin{equation}
      \mathcal{M} =  \delta_{ij} =  
    \begin{cases}
        True \quad \text{if } i=j\\
        False \quad \text{if } i \neq j\\ 
    \end{cases}
    \label{eq:mask}
\end{equation}
The vertex offsets are obtained as:
\begin{equation}
    \big\{ \boldsymbol{V}_\textrm{offset} \big\}^{1:T} = \mathcal{D}\Big(\boldsymbol{m}^{1:T} \, \big| \, \mathcal{T}, \mathcal{M}\Big) , 
    \label{eq:transformer_sequence}
\end{equation}
where $\mathcal{D}$ refers to the transformer decoder network. These predicted offsets $\big\{ \boldsymbol{V}_\textrm{offset} \big\}^{1:T}$ are added to the template mesh $\boldsymbol{T}$ to obtain mesh animation in canonical space as: 
\begin{equation}
    \big\{ \boldsymbol{V}_\textrm{pred} \big\}^{1:T} = \boldsymbol{T} + \big\{ \boldsymbol{V}_\textrm{offset} \big\}^{1:T} .
\end{equation}
The \textit{Expression2Latent MLP} $\mathcal{E}$ and the transformer decoder $\mathcal{D}$ are jointly trained with an L2-reconstruction loss:
\begin{equation}
    \mathcal{L}_\textrm{vertices} = \sum_{n=1}^{N} \Big( \sum_{t=1}^{T} \big|\big| \boldsymbol{V}_\textrm{gt}^{t} - \boldsymbol{V}_\textrm{pred}^{t} \big|\big|_2 \Big)_{n},
    \label{eq:seq_motion}
\end{equation}

The predicted vertices $\big\{ \boldsymbol{V}_\textrm{pred} \big\}^{1:T}$ are fed to our Optimized 3DGS avatar (Sec.~\ref{method:avatara_init}) and color related attributes of the avatar are further refined. We propose an alternating training strategy for the task as explained below. 

(a) In the first step, we predict vertex displacements (from the rest pose) in the canonical space for the entire sequence (Eq.~\ref{eq:transformer_sequence}). This  learns the optimal parameters for transformer $\mathcal{D}$ and Expression2Latent MLP $\mathcal{E}$ as:
\begin{equation}
    \mathcal{E}^{*}, \mathcal{D}^{*} = \argmin_{ \mathcal{E}, \mathcal{D} } \mathcal{L}_\textrm{vertices}  
\end{equation}

(b) In the second step, we predict the 3D Gaussian attributes with our Optimized 3DGS avatar (Sec.~\ref{method:avatara_init}) and render the full animation sequence.

The color MLP $\boldsymbol{\theta}_\textrm{color}$ of our optimized avatar is conditioned on predicted FLAME expression $\boldsymbol{e}_\textrm{pred}$ and per Gaussian latent $\boldsymbol{z}_i$, in addition to view direction $\boldsymbol{v}$, and predicts the view- and expression-dependent color as:
\begin{equation}
    \boldsymbol{c}_{i} = \boldsymbol{\theta}_\textrm{color} (\boldsymbol{e}_\textrm{pred}; \boldsymbol{z}_i; \boldsymbol{v} ).
\end{equation}
The predicted image $I_\textrm{pred}$ is obtained with the differentiable renderer $\mathcal{R}$ from Kerbl \etal~\cite{kerbl3Dgaussians} as:
\begin{equation}
    I_\textrm{pred} = \mathcal{R} \big(\{\boldsymbol{\mu_i}, \boldsymbol{s_i}, \boldsymbol{q_i}, \boldsymbol{c_i}\}^{1:G}, [R\,|\,t] \big) ,
\end{equation}
where $\boldsymbol{\mu_i}, \boldsymbol{s_i}, \boldsymbol{q_i}$ refers to the optimized avatar's position, scale, and rotations, respectively, and $G$ defines the total number of Gaussians.
The predictions are supervised with the photometric loss $\mathcal{L}_\textrm{photo}$ for the sequence:
\begin{equation}
    \mathcal{L}_\textrm{photo} = \sum_{t=1}^{T} \left( \mathcal{L}_\textrm{rgb} + \lambda_\textrm{g} \mathcal{L}_\textrm{global} + \lambda_\textrm{p} \mathcal{L}_\textrm{patch} \right)_t,
    \label{eq:seq_photo}
\end{equation}

In this step, we refine per-Gaussian latents $\boldsymbol{z}_{i}$ and Color MLP $\boldsymbol{\theta}_\textrm{color}$ with audio-conditioned expressions:
\begin{equation}
\boldsymbol{\theta}_\textrm{color}^{*}, \{\boldsymbol{z}_{i}^*\}^{1:G}  
    = \argmin_{\boldsymbol{\theta}_\textrm{color}, \{\boldsymbol{z}_{i}\}^{1:G}} \mathcal{L}_\textrm{photo}  
\end{equation}

Overall, we optimize two losses in the alternating fashion: (a) $\mathcal{L}_\textrm{vertices}$ which learns audio-conditioned facial motion and (b) $\mathcal{L}_\textrm{photo}$ which refines the optimized avatar for more accurate and photorealistic appearance. 
We do not refine the position, scale, rotation, and opacity; empirically, we found that they did not make a noticeable difference in the overall quality.

\begin{figure*}[h!]
    \centering
    \includegraphics[width=1.0\linewidth]{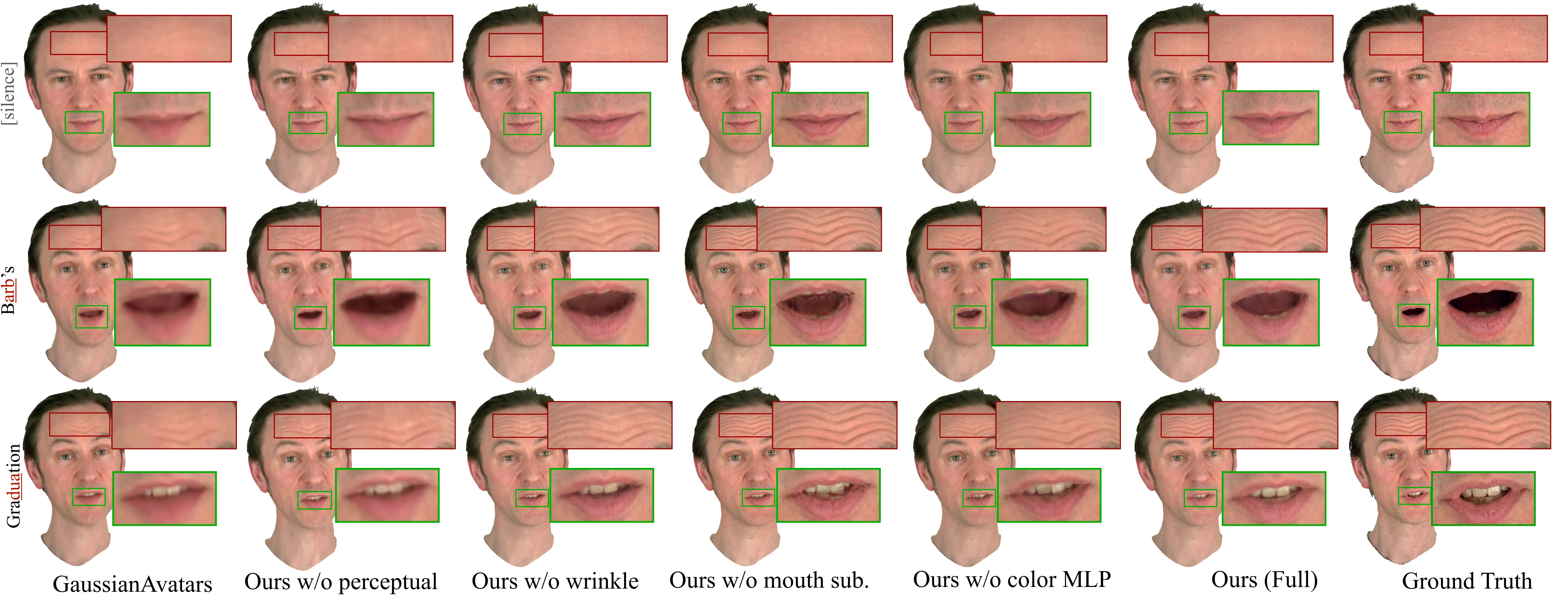}
    \vspace{-0.6cm}
    \caption{Avatar Reconstruction: GaussianAvatars~\cite{qian2023gaussianavatars} produces blurry results and cannot handle dynamic wrinkles. For our method, without perceptual loss it cannot synthesize sharp textures for global \& local  less observed regions like teeth, wrinkle regularization helps to model dynamic wrinkles, mouth faces subdivision helps with the better mouth interior and Color MLP helps synthesize sharper colors and accurate dynamic wrinkles. Our full avatar initialization technique with all regularization achieves the best results.}
    \label{fig:avatar_init}
    \vspace{1cm}
\end{figure*} 

\section{Results}\label{sec:results}
We evaluate \OURS{} on the tasks of (a) Avatar Representation and (b) Audio-Driven Animation. For (a), we evaluate standard perceptual image quality metrics SSIM, PSNR and LPIPS. For audio-driven animation, we evaluate lip synchronization LSE-D~\cite{wave2lip2020} as well as perceptual quality metrics. We train personalized avatars for different identities. Following GaussianAvatars~\cite{qian2023gaussianavatars}, we train on all 15 cameras except the frontal and report results on the frontal camera for all our experiments. All images are resized to $1604 \times 1100$ during training. For avatar reconstruction, we use 30 short sequences. For audio-driven animation, we use 300 sequences for training and 50 for val and test set each.
We encourage readers to watch the Supplementary Video for visual comparison of all results.

\subsection{Avatar Reconstruction}\label{results:avatar_init}
Compared to GaussianAvatars~\cite{qian2023gaussianavatars}, our proposed avatar initialization can generate high-quality results with as few as 30-35k points (see Fig.~\ref{fig:avatar_init} and Tab.~\ref{tab:avatar_initialization}). The perceptual loss helps increase the sharpness in the texture with fewer points. The wrinkle regularization helps to model dynamic wrinkles. Teeth subdivision helps with the better mouth interior. Color MLP helps synthesize sharper texture. Our full avatar initialization with all regularization achieves the best results.  We train our method on all except frontal camera and report results for the frontal camera. For these experiments, we show results for the most expressive actor from our dataset (Subject 4) and refer to Suppl. doc for others. 

\begin{table}[h!]
    \begin{center}
    \resizebox{1.0\linewidth}{!}{
    \begin{tabular}{l c c c c}
        \toprule
        Method & PSNR $\uparrow$ & SSIM $\uparrow$ &  LPIPS $\downarrow$ & {\# Gaussians} $\downarrow$ \\
        \toprule
        GaussianAvatar~\cite{qian2023gaussianavatars} & 26.53 & 0.9087  & 0.1487 & {98083}  \\
        \hline
        Ours (w/o perceptual) & 27.03  & 0.9116  & 0.1447 & \textbf{31875}  \\
        Ours (w/o wrinkle reg.)  & 28.10  & 0.9216  & 0.1312  &  33998 \\
        Ours (w/o mouth subdivision) & 28.35  & 0.9321   & 0.1244 &  34917  \\
        Ours (w/o Color MLP) & {28.93} & {0.9366}  &  {0.1235} & 32792   \\
        \hline
        Ours (Full) & \textbf{29.90} & \textbf{0.9495}  & \textbf{0.1104} & {32379}   \\
        \bottomrule
    \end{tabular}
    }
    \end{center}
    \vspace{-0.5cm}
    \caption{Avatar Reconstruction: With fewer Gaussian points, our method achieves superior quality compared to the alternate approaches. Perceptual loss increases the sharpness, wrinkle regularization models dynamic wrinkles, mouth subdivision learns better mouth interior, Color MLP synthesizes sharper colors and accurate dynamic wrinkles. The full avatar initialization with all regularizations achieves the best results.}\label{tab:avatar_initialization}
\end{table}

\begin{figure*}[t!]
    \centering
    \includegraphics[width=1.0\linewidth]{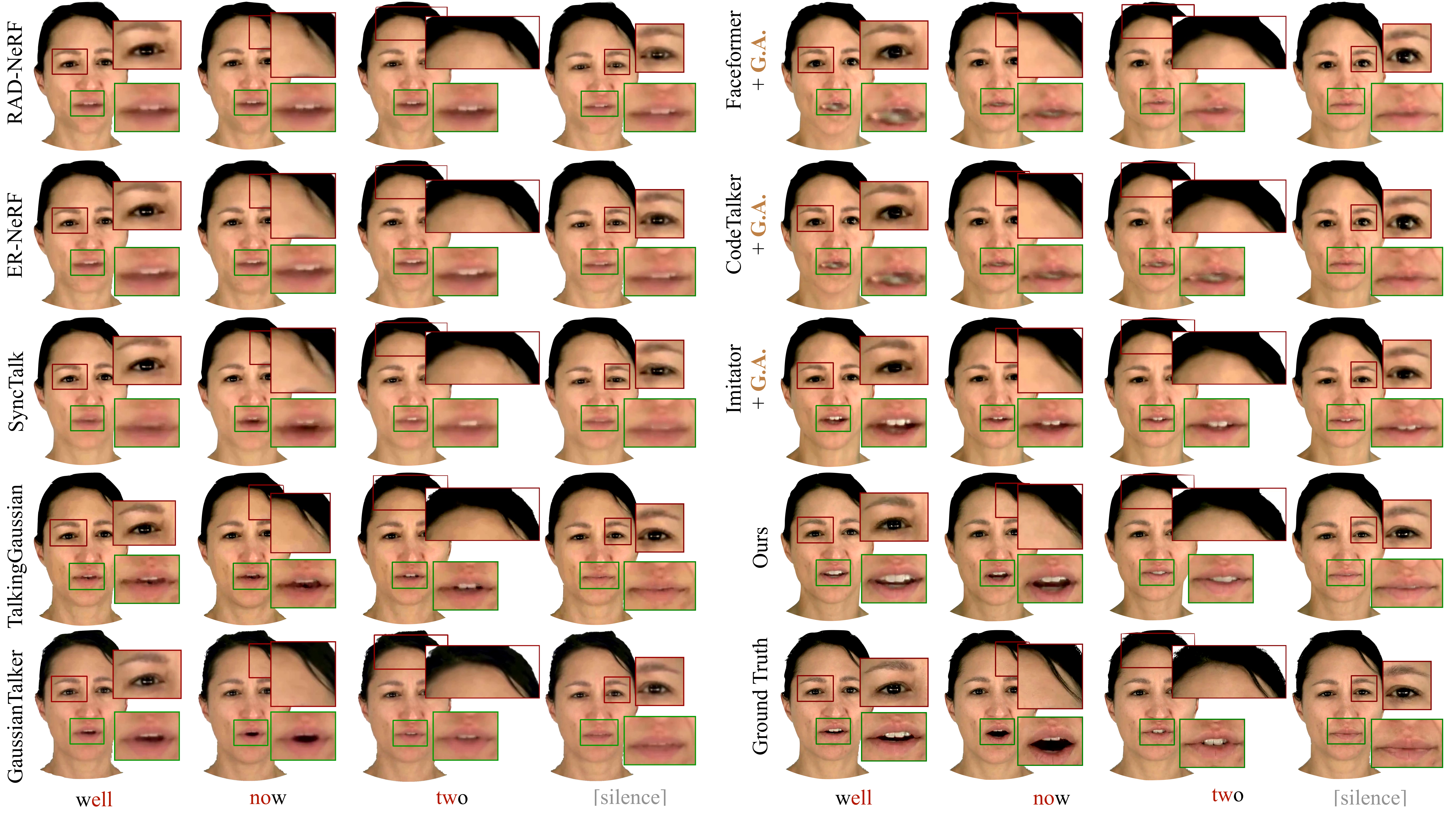}
    \vspace{-0.5cm}
    \caption{Baseline Comparison: We show comparisons against NeRF-based, 3DGS-based and FLAME animation methods combined with GaussianAvatars (\textcolor{brown}{\textbf{G.A.}})~\cite{qian2023gaussianavatars}. NeRF-based methods (RAD-NeRF~\cite{tang2022radnerf}, ER-NeRF~\cite{li2023ernerf} and SyncTalk~\cite{peng2023synctalk}) produce artifacts in texture as well as incorrect mouth articulations. 3DGS-based methods (TalkingGaussian~\cite{talkingGaussian2024} \& GaussianTalker~\cite{GaussianTalker2024}) can synthesize better lip-sync but produces blurry texture especially for mouth interior. Generalized FLAME animation methods (Faceformer~\cite{faceformer2022}, CodeTalker~\cite{xing2023codetalker}) show blurred mouth interiors, personalized methods (Imitator~\cite{Thambiraja_2023_ICCV}) produce better mouth interiors, however, the lip closures and synchronization is inaccurate. Our method outperforms all baselines both in lip-sync and photorealism.}
    \label{fig:baseline_comp}
\end{figure*}

\begin{table}[h!]
    \begin{center}
    \resizebox{1.0\linewidth}{!}{
    \begin{tabular}{p{0.2cm} l c c c c}
        \toprule
        & Method & LSE-D $\downarrow$ & PSNR $\uparrow$ & SSIM $\uparrow$ &  LPIPS $\downarrow$  \\
        \toprule
        \parbox[t]{1mm}{\multirow{3}{*}{\rotatebox[origin=c]{90}{{\small NeRF}}}}\,\,\ldelim\{{3}{*} & RAD-NeRF~\cite{tang2022radnerf}  & 13.17 & 13.15  & 0.8007  &  0.2741 \\
        & ER-NeRF~\cite{li2023ernerf} &  13.08& 15.94  &  0.8269 & 0.2512    \\
        & SyncTalk~\cite{peng2023synctalk}  & 12.50 & 18.24  & 0.8759  & 0.1920   \\
        \hline
        \parbox[t]{1mm}{\multirow{2}{*}{\rotatebox[origin=c]{90}{{\small 3DGS}}}}\,\,\ldelim\{{2}{*} & TalkingGaussian~\cite{talkingGaussian2024} & 12.38  &  20.29 & 0.8890  &  0.1745  \\
        & GaussianTalker~\cite{GaussianTalker2024}  & 12.19  & 20.32  & 0.8984  & 0.1724   \\
        \hline
        \parbox[t]{1mm}{\multirow{3}{*}{\rotatebox[origin=c]{90}{{\small FLAME}}}}\,\,\ldelim\{{3}{*} & Faceformer~\cite{faceformer2022} \textcolor{brown}{\textbf{+ G.A.}} & 11.86 &  22.18 & 0.9105  &  0.1608  \\
        & CodeTalker~\cite{xing2023codetalker} \textcolor{brown}{\textbf{+ G.A.}} & 11.68 & 22.23  & 0.9118  & 0.1595   \\
        & Imitator~\cite{Thambiraja_2023_ICCV} \textcolor{brown}{\textbf{+ G.A.}} & 11.61 & 22.83  & 0.9207  & 0.1519   \\
        \hline
        & Ours  & \textbf{11.25} & \textbf{24.73}  & \textbf{0.9362}  & \textbf{0.1286}  \\
        \bottomrule
    \end{tabular}
    }
    \end{center}
    \vspace{-0.5cm}
    \caption{Baseline Comparisons: we compare with NeRF-based, 3DGS-based and mesh-based (FLAME~\cite{flame_siggraphAsia2017}) baselines. We combine FLAME-based methods with 3DGS via GaussianAvatars (\textcolor{brown}{\textbf{G.A.}})~\cite{qian2023gaussianavatars}. Our method achieves superior results in both in perceptual quality as well as lip synchronization (LSE-D).}\label{tab:baseline_comparison}
\end{table}

\begin{figure*}[h!]
    \centering
    \includegraphics[width=\linewidth]{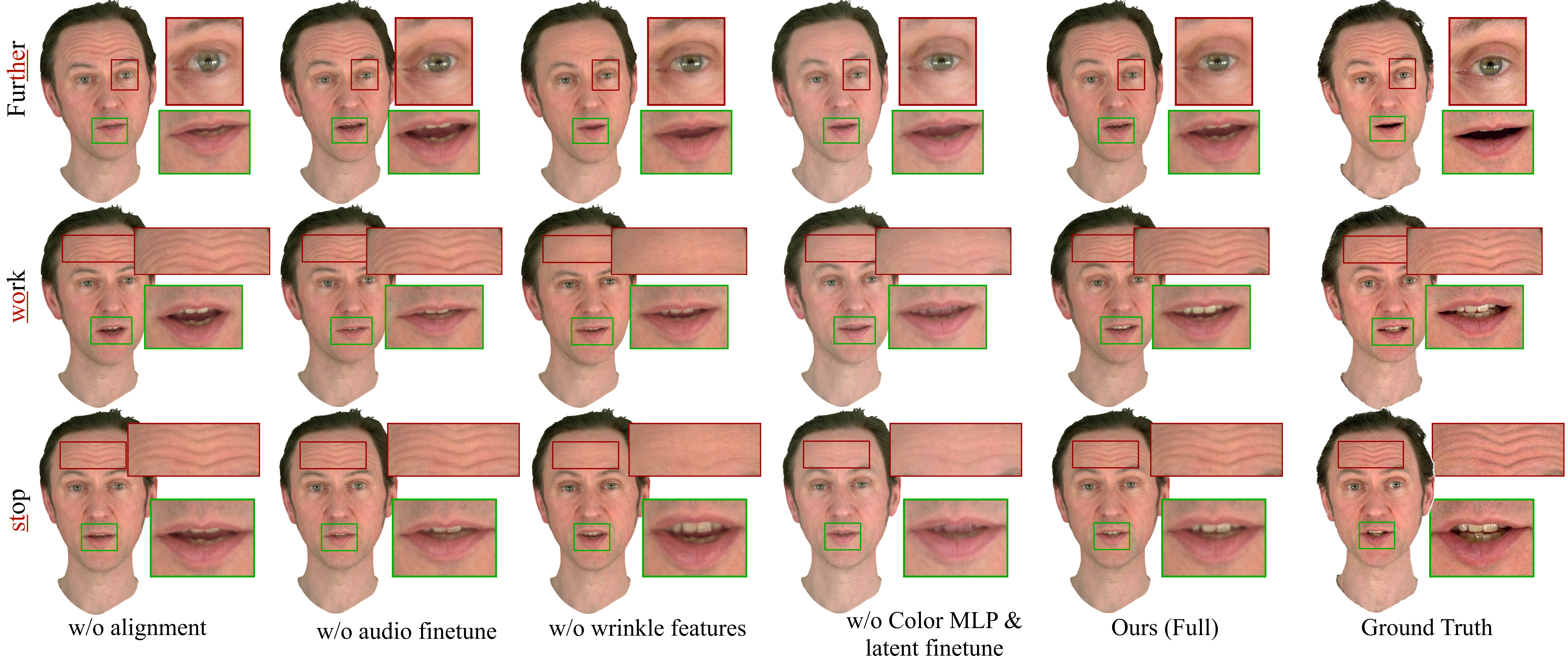}
    \vspace{-0.6cm}
    \caption{Ablation Study. Left-to-right: (1) Alignment mask is critical to properly infuse audio information into the sequence model. (2) Audio fine-tuning helps the method generate better lip sync. (3) Without wrinkle features, the model can not produce dynamic wrinkles. (4) Without fine-tuning Color MLP and latent features, the model produces bad mouth interiors and inaccurate dynamic wrinkles. Our full model with all the components achieves best results.
}
    \label{fig:method_ablation}
\end{figure*}

\subsection{Audio-Driven Animation}

\paragraph{Baseline Comparisons.} We compare our method against recent state-of-the-art methods. For NeRF- and 3DGS-based methods, we train on frontal camera since these methods are designed for monocular settings. There are no sequence models for audio-driven animation of 3D head avatars, thus, we combine audio-to-mesh animation methods~\cite{faceformer2022, xing2023codetalker, Thambiraja_2023_ICCV} with current state-of-the-art mesh-to-3D avatar creation method~\cite{qian2023gaussianavatars}. We report results on the front camera for fairness, since some methods are designed only for front/single camera only. We report results averaged over all subjects, see Fig.~\ref{fig:baseline_comp} and Tab.~\ref{tab:baseline_comparison}. Our method consistently achieves better results than baselines both in terms of perceptual quality and lip synchronization.

\begin{table}[h!]
    \begin{center}
    \resizebox{1.0\linewidth}{!}{
    \begin{tabular}{l c c c c}
        \toprule
        Method & LSE-D $\downarrow$ & PSNR $\uparrow$ & SSIM $\uparrow$ &  LPIPS $\downarrow$  \\
        \toprule
        w/o alignment  & 12.66 & 21.02 & 0.9104 & 0.1855  \\
        w/o audio finetune  & 11.78 & 22.73  & 0.9355 &  0.1198 \\ 
        w/o wrinkle features & 11.28 & 23.14 & 0.9311  & 0.1162   \\
        w/o color MLP \& latent finetune & 11.32 & 23.96 & 0.9367 & 0.1133  \\
        \bottomrule
        Ours (Full)  & \textbf{11.15} & \textbf{24.97} & \textbf{0.9470}  & \textbf{0.1101}   \\
        \bottomrule
    \end{tabular}
    }
    \end{center}
    \vspace{-0.5cm}
    \caption{Ablation study. Without alignment mask, the model ignores the audio signal. Audio fine-tuning helps to improve lip sync. Wrinkle features help with dynamic wrinkles and overall realism. Finetuning Color MLP and latents rectifies the inaccurate mouth interior. Our full model achieves the best results.
} \label{tab:ablation_study}
\end{table}

\vspace{-0.7cm}
\paragraph{Ablation Study.} Finally, we ablate different design choices of our method on most expressive actor from our dataset (Subject 4) in Fig.~\ref{fig:method_ablation} and Tab.~\ref{tab:ablation_study}. Alignment mask is critical for accurately infusing audio features into the sequence model. Without audio fine-tuning refers to using generic audio features without any personalization of lip encoder, without audio model fine-tuning the model produces incorrect lip synchronization. Without wrinkle features refers to setting without using wrinkle features for producing FLAME expressions. Without wrinkle features the method cannot produce dynamic wrinkles. Without finetuning Color MLP \& latent features with predicted expressions from our Expression encoder, the method produces bad mouth interiors and inaccurate dynamic wrinkles. Our full model with all components achieves best results. 
We refer readers to supplemental video for visual comparison.

 \section{Conclusion}
In this work, we propose a novel approach to create high-fidelity and photorealistic 3D head avatars that can be animated from audio input. We designed the first transformer-based sequence model 
for audio-driven head animation of 3DGS based avatar. Our sequence model is made possible by a lightweight and compact avatar initialization based on 3D Gaussian Splatting. We proposed several regularization techniques to handle dynamic wrinkles, skin creasing and sharpness of the texture. Our method produces (a) photorealistic and high-quality 3D head avatars that can be rendered from arbitrary viewpoints (b) visually natural animations like skin creasing during talking. We believe
this is an important first step towards enabling the animation
of detailed and lightweight 3D head avatar, which can enable many new possibilities for content creation and digital avatars for immersive telepresence.

\section{Acknowledgments}
\vspace{-0.2cm}
This work was supported by the ERC Starting Grant Scan2CAD (804724), the Bavarian State Ministry of Science and the Arts and coordinated by the Bavarian Research Institute for Digital Transformation (bidt), the German Research Foundation (DFG) Grant ``Making Machine Learning on Static and Dynamic 3D Data Practical,'' the German Research Foundation (DFG) Research Unit ``Learning and Simulation in Visual Computing''. We would like to thank Shenhan Qian for help with tracking. 
{
    \small
    \bibliographystyle{ieeenat_fullname}
    \bibliography{main}
}

\setcounter{section}{0}

\clearpage
\maketitlesupplementary
\appendix

\renewcommand\thesection{\Alph{section}}

In this supplemental document, we provide additional results of our proposed method GaussianSpeech, including a user study, in Sec.~\ref{sec:additional_results}.
Details about our network architecture and training scheme are given in Sec.~\ref{sec:architecture}, including preliminaries used in the main paper, see Sec.~\ref{sec:prelimiaries}.
In Sec.~\ref{sec:baselines}, we have added a further discussion of the baseline methods, and in Sec.~\ref{sec:dataset_details} we provide more dataset details.

\section{Additional Experiments}\label{sec:additional_results}

\subsection{Novel View Synthesis}
\begin{figure}[h]
    \centering
    \includegraphics[width=0.9\linewidth]{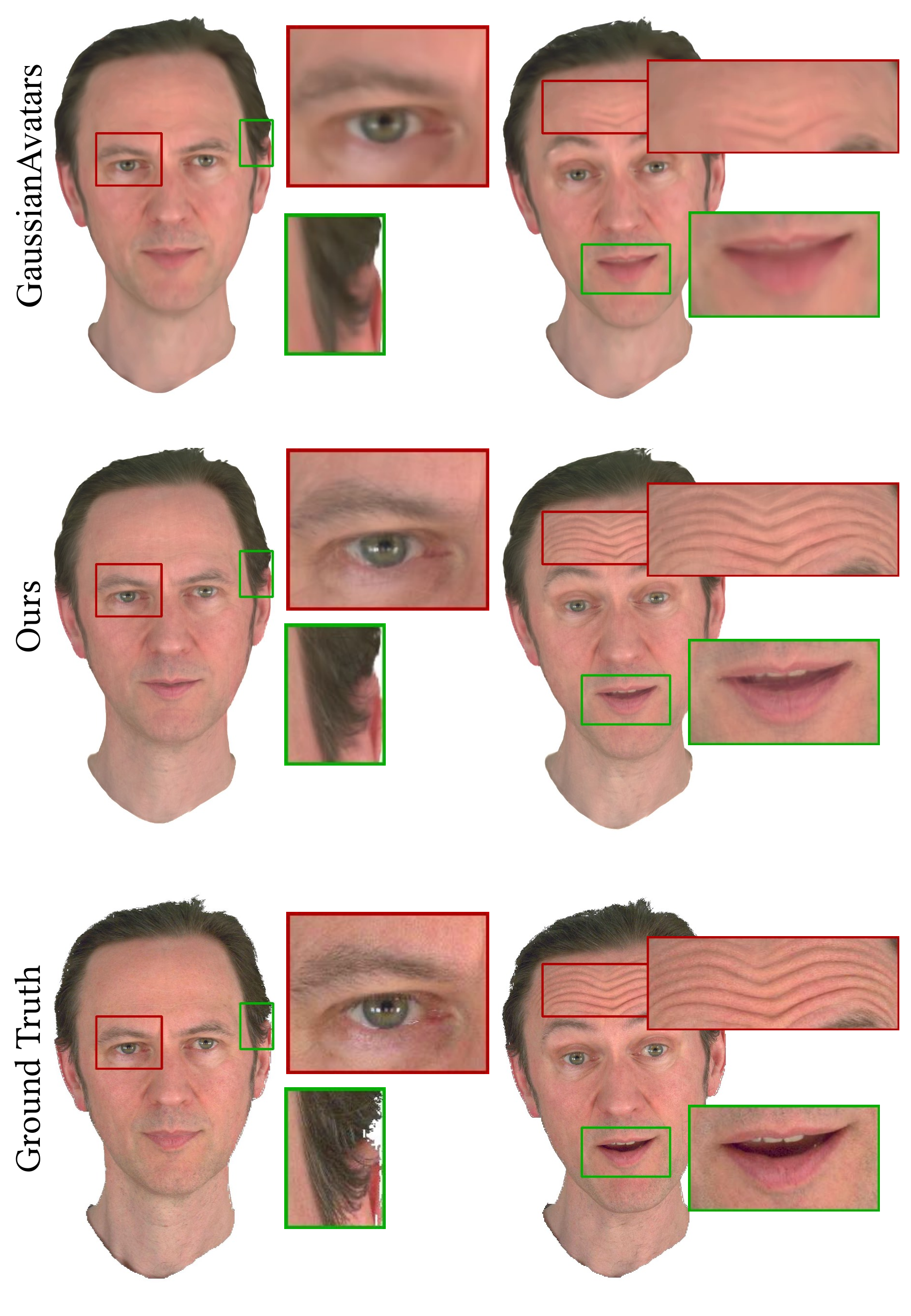}
\caption{Novel View Synthesis quality in comparison to\linebreak GaussianAvatars~\cite{qian2023gaussianavatars}. In contrast to ours, GaussianAvatars generates blurry texture and cannot generate dynamic wrinkles.}
    \label{fig:nvs_mike}
\end{figure}

\begin{figure*}[h!]
    \centering
    \includegraphics[width=0.9\linewidth]{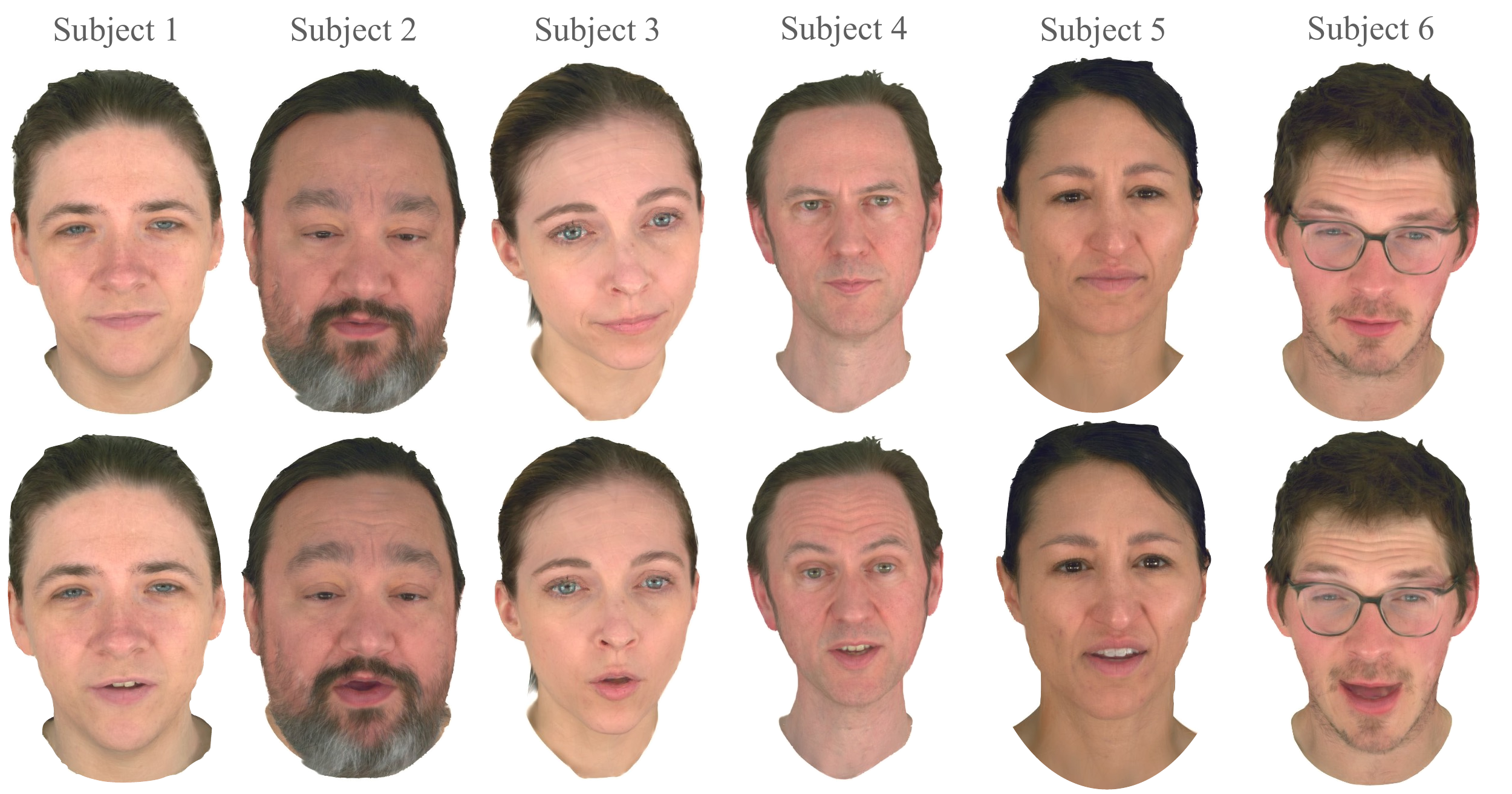}
    \caption{Reconstructed Avatars: We show the reconstructed avatars from novel views for all the participants from our dataset. We visualize two randomly selected frames for each avatar, one where the avatar is silent and the other where the avatar is speaking. Our method generates high quality avatars generating realistic and sharper textures.}
    \label{fig:data_avatars}
\end{figure*}

\begin{figure*}[h!]
     \centering
     \includegraphics[width=1.0\linewidth]{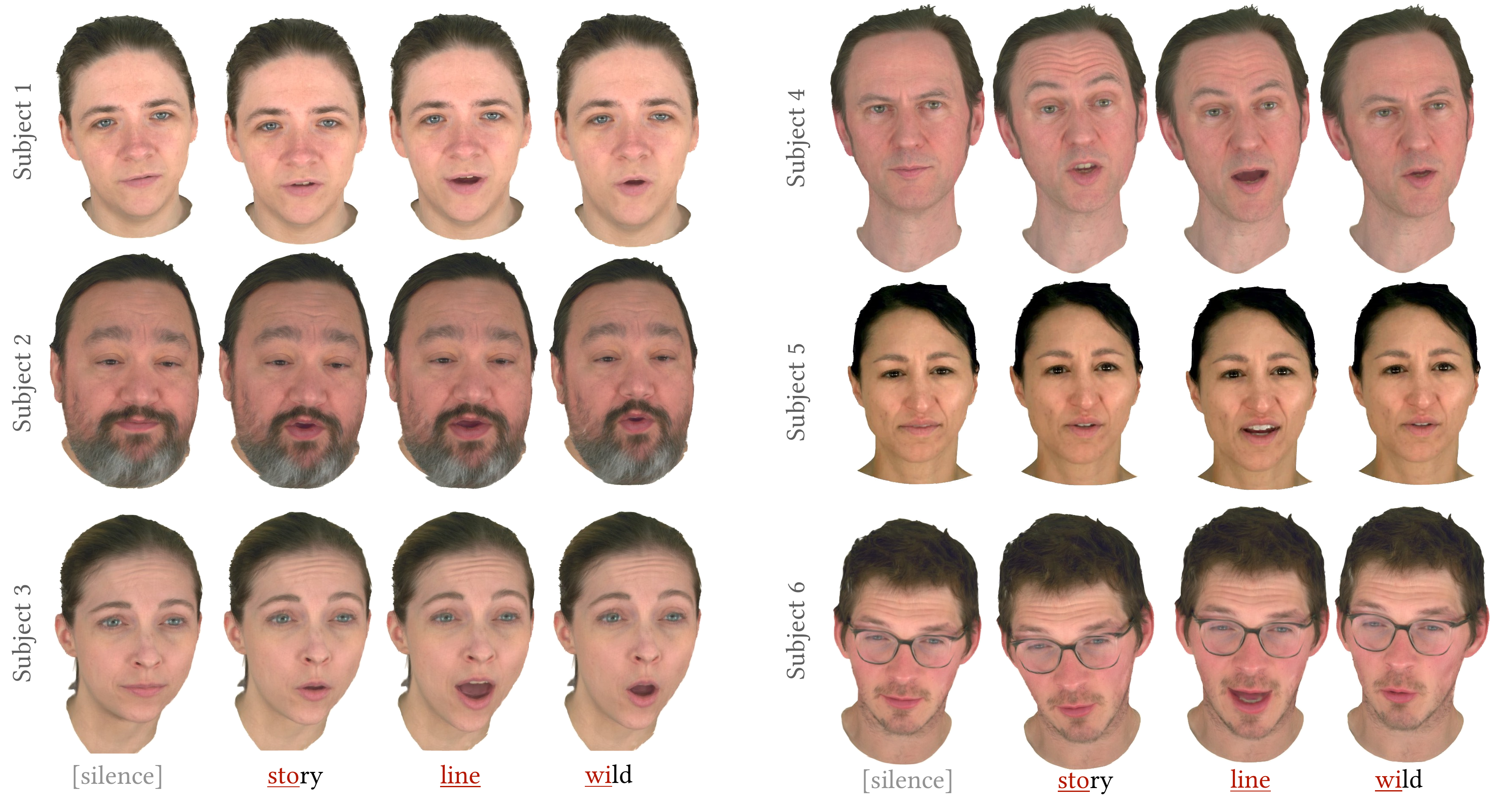}
     \caption{Audio-Driven Animation: We show animation results of our avatars animated directly from audio signal for the novel camera. The words spoken by the avatars are highlighted at the bottom. }
     \label{fig:audio_avatars}
 \end{figure*}

In our  experiments, we found that training with at least 30 sequences enables generalizing for mouth articulations.
We show novel view synthesis results and zoom-ins in Fig.~\ref{fig:nvs_mike}. 
For all the avatars from our dataset, we show results for avatar initialization in Fig.~\ref{fig:data_avatars} and Tab.~\ref{tab:avatar_init_points}.
Audio-driven animations are shown in Fig.~\ref{fig:audio_avatars}.

\vspace{4cm}
\begin{table}[h!]

     \begin{center}
    \resizebox{1.0\linewidth}{!}{
    \begin{tabular}{l c c c c}
        \toprule
        Subjects & PSNR $\uparrow$ & SSIM $\uparrow$ &  LPIPS $\downarrow$ & {\# Gaussians} $\downarrow$ \\
        \toprule
        Subject 1 & 30.07  & 0.9598  & 0.0754  & 30653  \\  Subject 2 & 30.13  & 0.9180  & 0.1125  & 32443  \\  Subject 3 & 28.62  & 0.9607  & 0.1072  & 31434  \\  Subject 4 & 29.90  & 0.9495  & 0.1104  & 32379  \\  Subject 5 & 30.05  & 0.9529 & 0.1185  & 29490  \\  Subject 6 & 26.82  & 0.9228  & 0.1322  & 35138  \\  \bottomrule
    \end{tabular}
    }
    \end{center}
    \vspace{-0.55cm}
     \caption{Avatar Initialization: All avatars converge in the range of 29-35K Gaussians. We show perceptual quality metrics for each of our avatars evaluated for novel views.}
     \label{tab:avatar_init_points}
\end{table}

\subsection{Effect of Latent Features}
\begin{figure}[h!]
    \centering
    \includegraphics[width=0.8\linewidth]{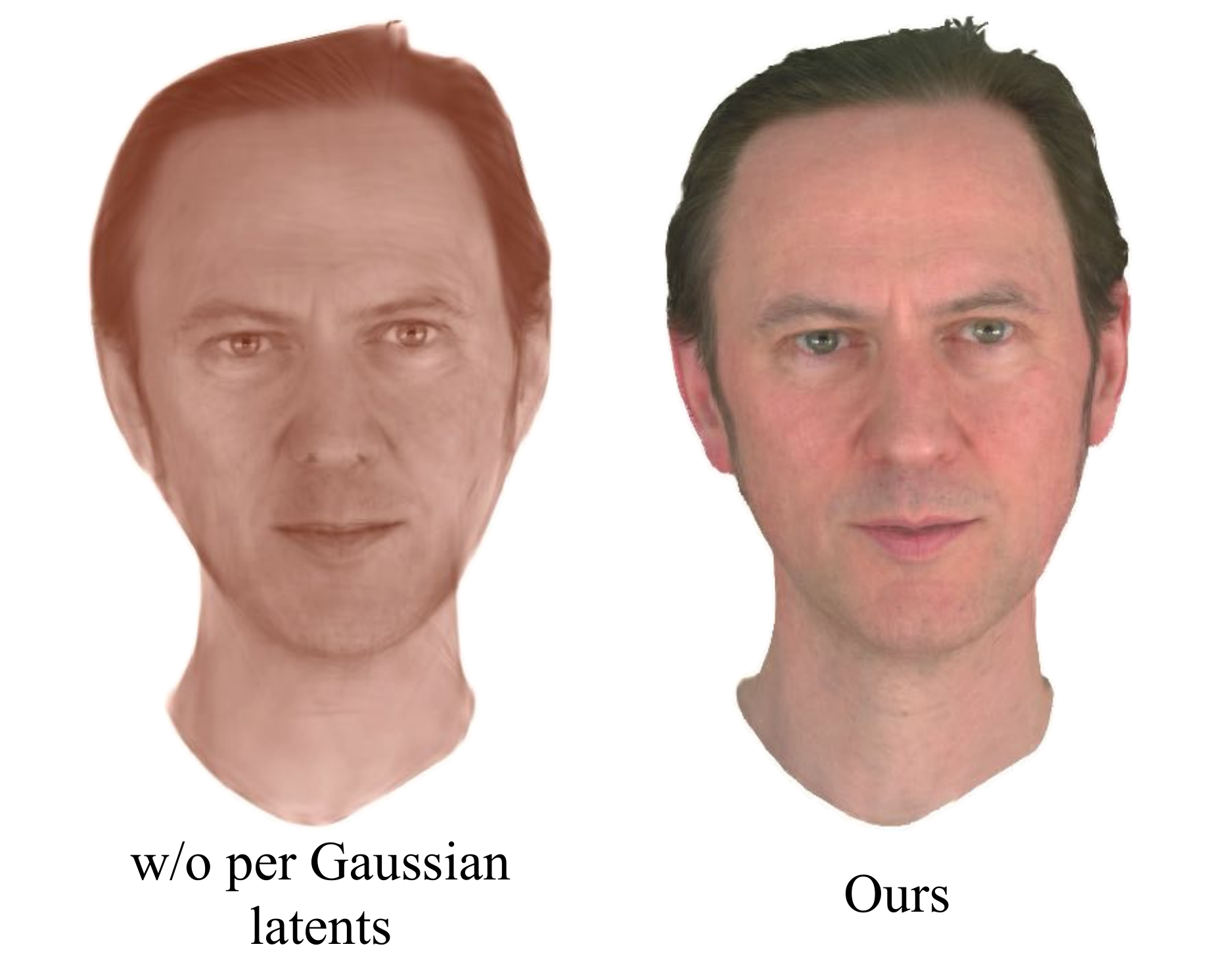}
    \caption{Effect of per Gaussian latents: Without the features, the color MLP cannot produce accurate texture colors. }
    \label{fig:nvs_per_gaussian_latents}
\end{figure}

We analyze the effect of per Gaussian latent features during our avatar initialization stage in Fig.~\ref{fig:nvs_per_gaussian_latents}. Note that the per Gaussian features are critical to produce accurate texture colors for the avatar.

\subsection{Failure Cases}
\begin{figure}[h!]
     \centering
     \includegraphics[width=1.0\linewidth]{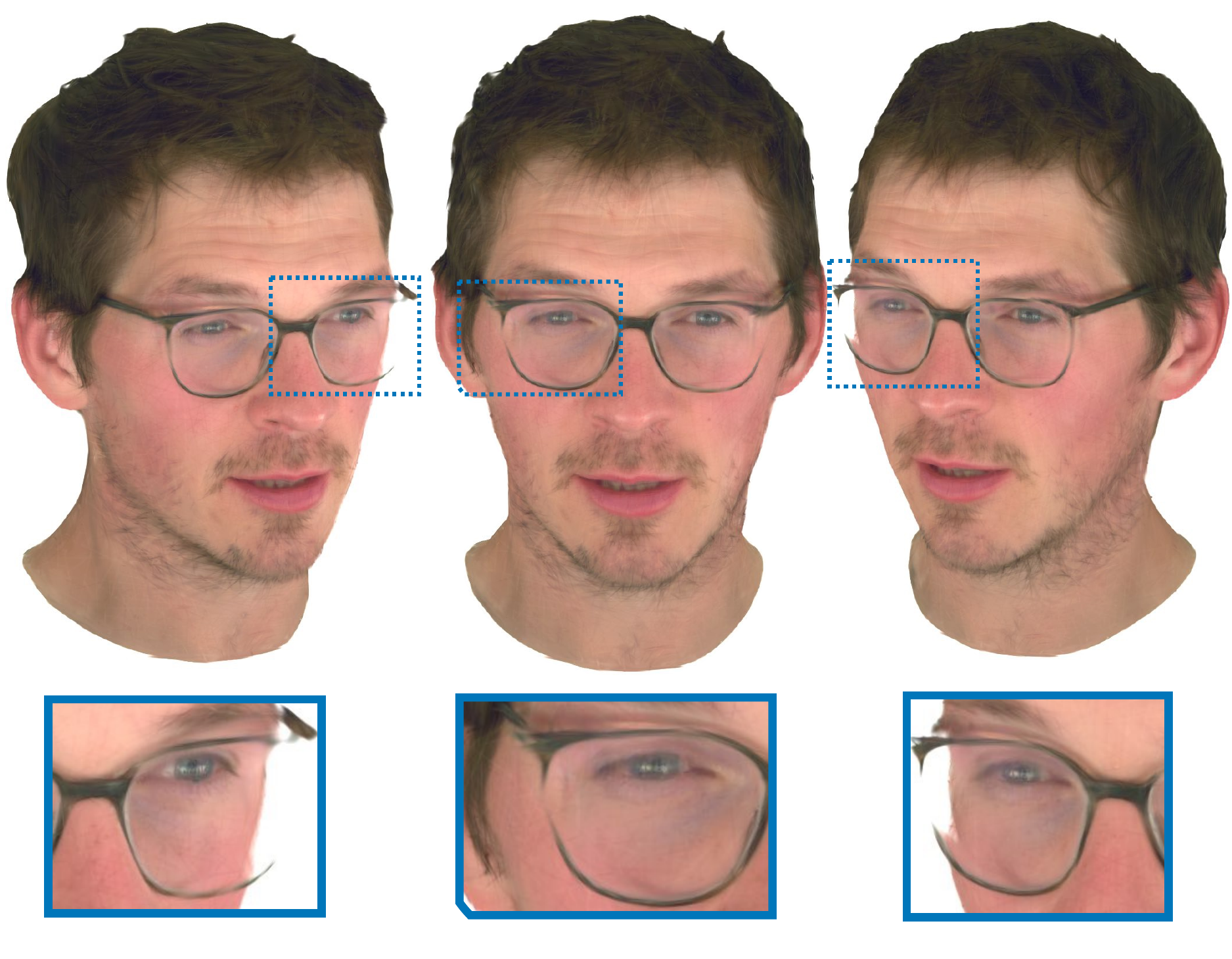}
    \vspace{-0.3cm}
     \caption{ Failure Cases: The method fails to produce accurate accurate glass
geometry and specularities on the surface of glasses.}
     \label{fig:failure}
 \end{figure}

While our method produces photorealistic and high-quality animations in synchronization with audio, it also has several limitations. Our avatar initialization strategy is based on FLAME~\cite{flame_siggraphAsia2017}; thus, our method struggles with avatars wearing accessories like glasses. The glass geometry and specularities on the surface of glasses can not be accurately produced and fails during free-viewpoint rendering, see Fig.~\ref{fig:failure}. 
In the future, this can be improved by designing better models for representing human head geometry instead of 3D mesh. 
Also, our texture representation based on the Color MLP has baked-in lighting, and cannot be separated from material properties, which is important for placing avatars in different environments (e.g., during immersive telepresence). 

\subsection{User Study}
\begin{figure}[h!]
  \centering
  \includegraphics[width=1.0\linewidth]{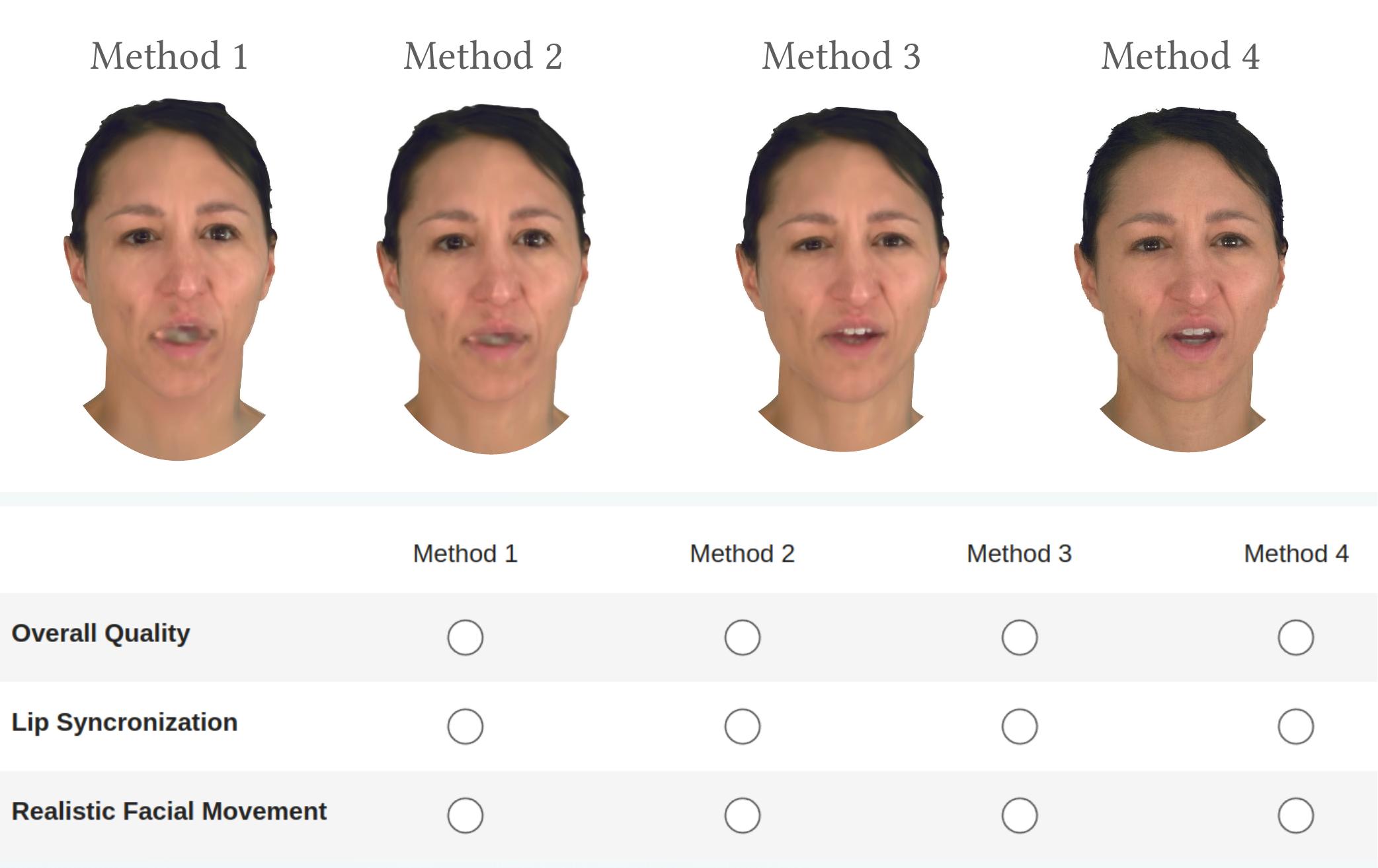}
  \vspace{-0.7cm}
  \caption{Different methods shown to users and questions asked to assess the quality of different methods during perceptual study evaluation. Method names were anonymized to avoid bias towards a particular method.  Users were asked to select one of the shown methods for each of the three questions based on which one they believe exhibits best results.}
  \label{fig:user_study_questions}
\end{figure}

To evaluate the fidelity based on human perceptual evaluation, we performed a user study with 30 participants over a set of 15 questions. The users were given a carefully crafted set of instructions to evaluate (a) Overall Animation Quality (b) Lip Synchronization and (c) Realism in Facial Movements. 
The users were asked to assess different anonymous methods (including \OURS{}) on these three parameters.
In the course of the study, participants were presented with these questions to focus on different aspects of 3D facial animation, shown in Fig~\ref{fig:user_study_questions}. For every question, participants were instructed to meticulously evaluate the provided methods and select the option that best aligned with their judgment.

\begin{figure}[h!]
  \centering
  \includegraphics[width=1.0\linewidth]{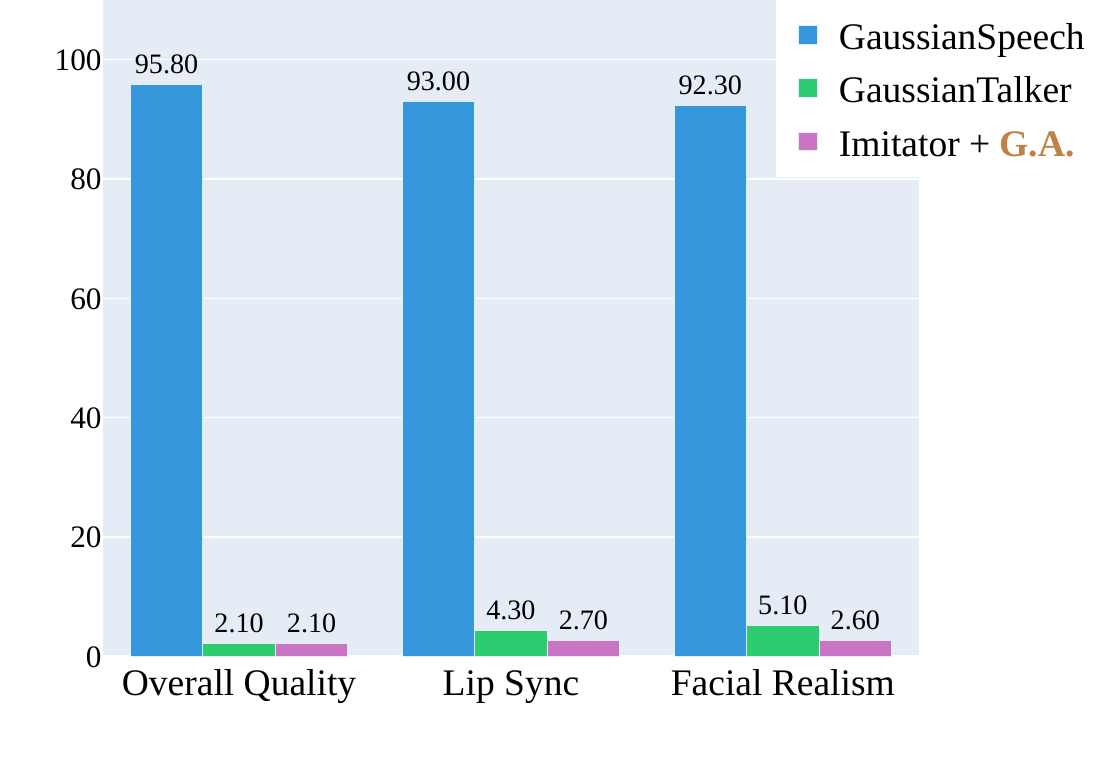}
  \vspace{-0.8cm}
  \caption{User study comparison with baselines. We measure preference for (1) Overall Animation Quality, (2) Lip Synchronization and (3) Facial Realism. \OURS{} results are overwhelmingly preferred over the best baseline methods on all these aspects. }
  \label{fig:user_study_vid}
\end{figure}

For the first question evaluating overall quality, participants were instructed to consider factors such as visual appeal, clarity, and general impression, and to choose the method number that they believed demonstrates the highest overall quality.
For the second question, participants were directed to evaluate the lip synchronization of each animation method. They were prompted to pay close attention to how well the lip movements aligned with the spoken words or sounds. Participants were reminded to select only one option that, in their judgment, exhibited the best lip synchronization.
Lastly, the third question was focussed on evaluating the realistic facial movement of each 3D facial animation method. Participants were instructed to consider the naturalness and persuasiveness of facial expressions and movements and to choose the method number that, in their opinion, demonstrates the most realistic facial movement. Again, participants were reminded to select only one option per question throughout the study.

Our method consistently achieves better lip-audio synchronization while also representing fine-scale
facial details like skin creasing and wider
mouth motions. This is confirmed by our perceptual user
study in Fig.~\ref{fig:user_study_vid}.

\subsection{Inference Speed}
We report inference speed averaged over the test set on a single Nvidia RTX 2080 Ti with 12GB VRAM as well as NVIDIA RTX A6000 with 48GB VRAM. Since TalkingGaussian~\cite{talkingGaussian2024} network does not fit in 12 GB VRAM; we report its inference time only for a 48 GB VRAM (NVIDIA A6000). The results are presented in Tab.~\ref{tab:inference_speed}.

\begin{table}[h!]

     \begin{center}
     \resizebox{1.0\linewidth}{!}{
     \begin{tabular}{c c c c}
         \toprule
         Method &  {FPS (2080Ti)}$\uparrow$ & {FPS (A6000)}$\uparrow$ & \# Gaussians \\
         \toprule
         Faceformer~\cite{faceformer2022} \textcolor{brown}{\textbf{+ G.A.}} & 25.38 & 42.23 & 65-100K \\
         CodeTalker~\cite{xing2023codetalker}  \textcolor{brown}{\textbf{+ G.A.}}  & 23.92 & 38.98 & 65-100K \\
         Imitator~\cite{Thambiraja_2023_ICCV} \textcolor{brown}{\textbf{+ G.A.}} & 23.32 & 39.71  &  65-100K  \\
         SyncTalk~\cite{peng2023synctalk} & 10.14 & 21.51 & N/A  \\
         ER-NeRF~\cite{li2023ernerf} & 16.88 & 17.98 &  N/A   \\
         RAD-NeRF~\cite{tang2022radnerf} & 17.18 & 21.41 &  N/A   \\  
         GaussianTalker~\cite{GaussianTalker2024} & 57.82 & 59.01 & 41K-44K  \\
         TalkingGaussian~\cite{talkingGaussian2024} & OOM  &  73.34  & 31K-98K  \\
         Ours  & \textbf{74.29} & \textbf{123.48}  &  30-35K    \\   
     \bottomrule
     \end{tabular}
     }
     \end{center}
     \vspace{-0.55cm}
      \caption{Inference Speed on Nvidia RTX 2080 Ti (12GB VRAM) and NVIDIA A6000 (48GB VRAM). TalkingGaussian's network does not fit in 12 GB VRAM and throws Out-of-Memory (OOM) error; for this method, the inference time is only given for a 48 GB VRAM (NVIDIA A6000). 
}
     \label{tab:inference_speed}
 \end{table}

\begin{figure}[h!]
  \centering
  \includegraphics[width=0.8\linewidth]{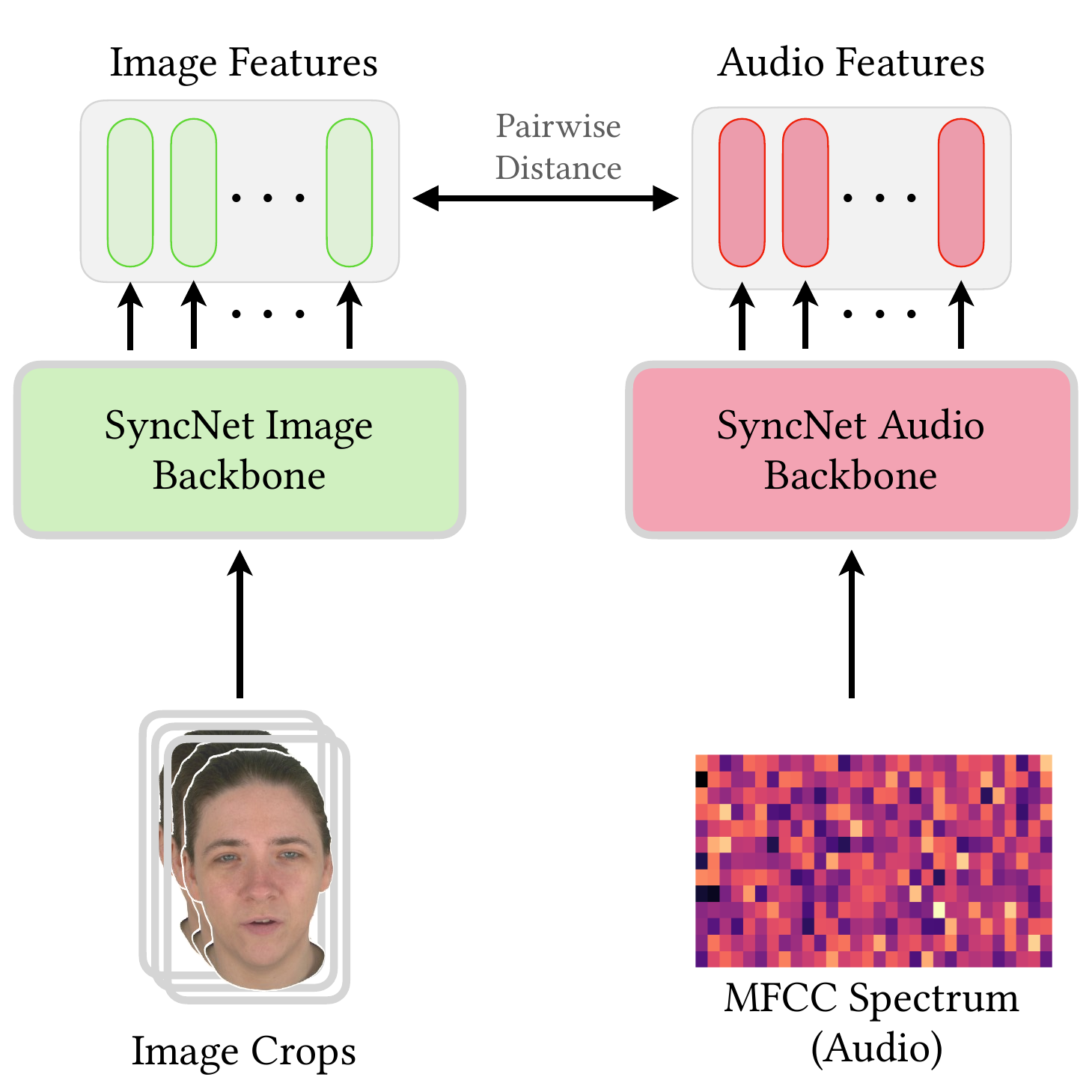}
  \vspace{-0.2cm}
  \caption{The image crops are passed to the pretrained Syncnet image backbone to extract image features, and audio features, represented as MFCC power spectrum, are extracted via pretrained Syncnet audio backbone. Finally, the pairwise distance between image and audio features are calculated to compute lip synchronization.}
  \label{fig:lsed_d_eval}
\end{figure}

\section{Architecture \& Training Details}\label{sec:architecture}

\OURS{} is implemented using PyTorch Lightning framework~\cite{falcon2019pytorch} with Wandb~\cite{wandb} for logging.

\subsection{Implementation Details}

\paragraph{Avatar Representation.} During avatar initialization, the teeth subdivision changes the initial mesh with 5143 vertices and 10144 faces to 5431 vertices and 10648 faces. For perceptual loss $\mathcal{L}_\textrm{global}$, we downscale images to $401 \times 275$. The volume based pruning ensures that the avatar converges to 30-35k Gaussian points. We use Adam optimizer with exponential decay and the default learning rates from ~\cite{qian2023gaussianavatars}.  For the loss (Eq.~\ref{eq:avatar_equation} main paper), we use $\lambda_\textrm{pos}=0.01$, $\lambda_\textrm{s}=1$, $\lambda_\textrm{g}=1$, $\lambda_\textrm{p}=0.001$ and $\lambda_\textrm{w}=10$. For rendering, we use the differentiable tile rasterizer~\cite{kerbl3Dgaussians}.

\paragraph{Audio-Driven Sequence Generation.} For the first 10K iterations, we train the Lip transformer, Wrinkle transformer and Expression encoder each, and then use them in our training pipeline. For the vertex loss $\mathcal{L}_\textrm{vertices}$ (Eq.~\ref{eq:seq_motion} main paper), we predict 5431 offsets obtained after mouth subdivision. For the next 2000 iterations, we train the model only with $\mathcal{L}_\textrm{vertices}$ to ensure that it learns coarse motion reasonably in coherence with the audio. And then gradually add $\mathcal{L}_\textrm{photo}$, and train the model together with $\mathcal{L}_\textrm{vertices}$ and $\mathcal{L}_\textrm{photo}$ in an alternating fashion.

\subsection{Architecture Details}
\begin{figure}[h!]
  \centering
  \includegraphics[width=1.0\linewidth]{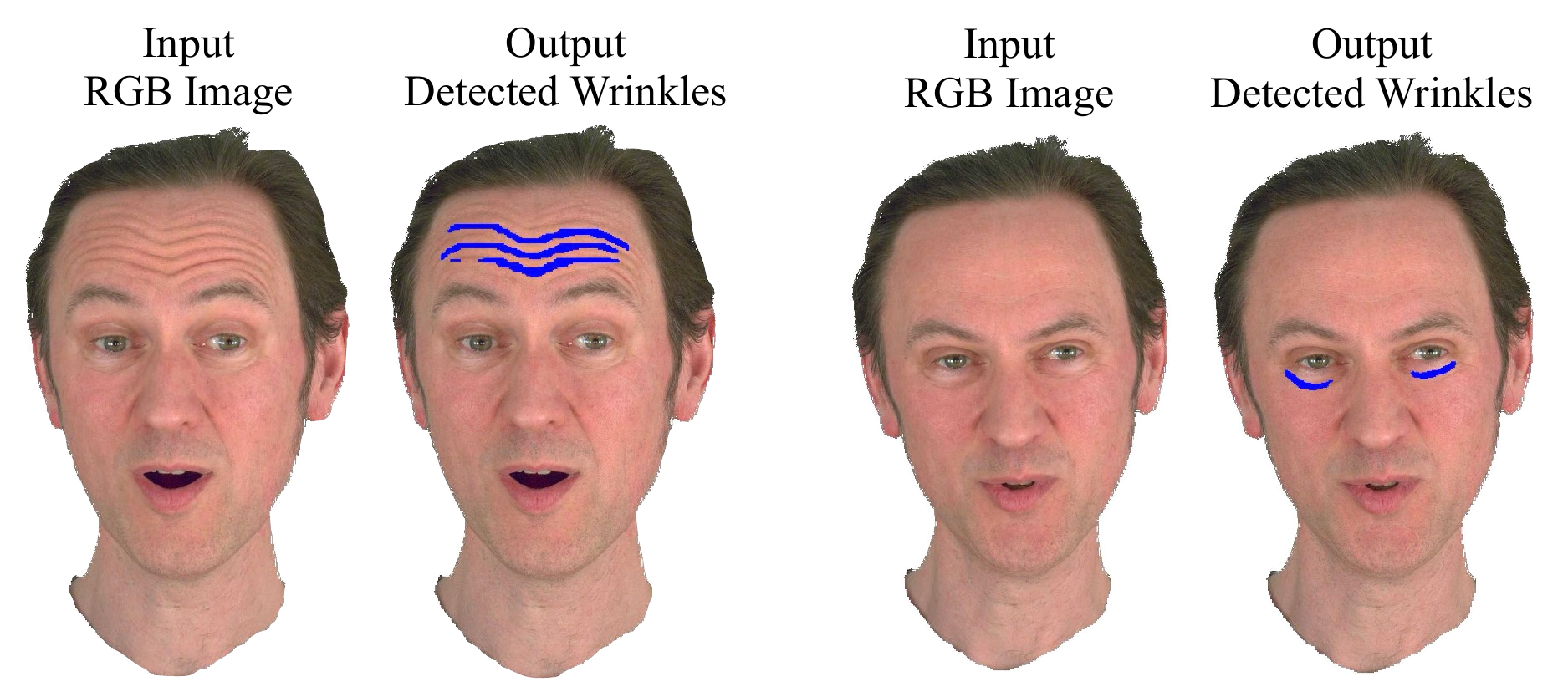}
  \vspace{-0.5cm}
  \caption{Sample output of the wrinkle detection model~\cite{wrinkle_detection} used for wrinkle regularization loss during avatar training.}
  \label{fig:sample_wrinkles}
\end{figure}

\textbf{Avatar Initialization.} During avatar initialization (Sec.~\ref{results:avatar_init}, main paper), we randomly sample 16 random patches per iteration with a patch size of $128 \times 128$ from the facial area. We start with an initial learning rate of 5e-3 and exponentially decay until 5e-5. We perform densification every 5000 iterations and we do not reset opacity. We train with the batch size of 1 with Adam optimizer, render images on white background and train for 100,000 iterations on RTX 2080 Ti (12 GB VRAM). Our avatars converge between 29-35K Gaussian points, as shown in Tab.~\ref{tab:avatar_init_points}. We show sample output of our wrinkle detection model used in main paper in Fig.~\ref{fig:sample_wrinkles}.

\noindent
\textbf{Audio-to-Avatar Model.} 
For the audio encoder, the TCN layers of the Wav2Vec 2.0~\cite{baevski2020wav2vec} are initialized with the pre-trained wav2vec 2.0 weights trained on a large corpus of audio data from different languages and is frozen during fine-tuning. The Frequency Interpolation layer simply performs linear interpolation of the incoming features and has no learnable parameters. 
For the Lip an Wrinkle encoder, we use latent dimension of 64 and for Expression encoder and the Transformer decoder the latent dimension is 128. For the multihead self and cross attention layers of the transformer decoder, we use 4 heads and set the dimension to 1024 for each decoder block. We use an Adam optimizer with a learning rate of 1e-4 and update the model one sequence per iteration, train on Nvidia RTX A6000 (48 GB VRAM) for 100,000 iterations.

\noindent
\textbf{Evaluation Metrics.} To evaluate lip synchronization of the generated mouth expressions with the audio signal, we use LSE-D (Lip Sync Error Distance)~\cite{wave2lip2020}. Specifically, this involves feeding rendered face crops and the corresponding audio signal into a pre-trained SyncNet~\cite{Chung16a} to evaluate how close the acoustic signal matches the phonetic movements. The facial movements are encoded as crops of only the facial region, and the audio signal is represented as MFCC power spectrum. These are then passed into the pretrained SyncNet backbone~\cite{Chung16a} and the pairwise distance is evaluated, as shown in Fig.~\ref{fig:lsed_d_eval}.

\section{Preliminaries}\label{sec:prelimiaries}

\subsection{3D Gaussian Splatting}

Recently, 3D Gaussian Splatting~\cite{kerbl3Dgaussians} has emerged as a promising approach to represent a static scene explicitly with anisotropic 3D gaussian directly from multiview images and estimated/given camera poses. Specifically, it represents a scene using a set of 3D Gaussian splats, each defined by a set of optimizable parameters, including a mean position $\boldsymbol{\mu} \in \mathbb{R}^{3} $ and a positive semi-definite covariance matrix $\Sigma \in \mathbb{R}^{3 \times 3}$ as:

\begin{equation}
    G(\boldsymbol{x}) = e^{- \frac{1}{2} (\boldsymbol{x}-\boldsymbol{\mu})^T \Sigma^{-1} (\boldsymbol{x}-\boldsymbol{\mu})   }
\end{equation}

Given that covariance matrix $\Sigma$ needs to be positive semidefinite to have physical meaning and gradient-based optimization methods cannot be constrained to produce such valid matrices, Kerbl \etal~\cite{kerbl3Dgaussians} first define an ellipsoid with scaling matrix $S$ and rotation matrix $R$ as:
\begin{equation}
    \Sigma = RSS^{T}R^{T}.
\end{equation}

To allow for independent optimization for scale and rotation, separate 3D vectors are stored. The scale is represented using a scaling vector $\boldsymbol{s} \in \mathbb{R}^3$ and a quaternion $\boldsymbol{q} \in \mathbb{R}^4$ for rotation. 

For rendering every pixel on the image, the color $\boldsymbol{C}$ is computed by blending all the 3D Gaussians overlapping a pixels as:

\begin{equation}
    \boldsymbol{C} = \sum_{i=1}^{N} \boldsymbol{c}_i \alpha_i \prod_{j=1}^{i-1} (1-\alpha_j)
\end{equation}
where $\boldsymbol{c}_i$ refers to 3-degree spherical harmonics (SH) ~\cite{sph_irradiance} color obtained by blending $N$ ordered points overlapping the pixel, blending weight $\alpha_i$ is given by multiplying 2D projection of the 3D Gaussian with learnt per-point opacity. Paired with a differentiable tile rasterizer, this enables real-time rendering. To handle complex scenes and respect visibility order, depth-based sorting is applied to the Gaussian splats before blending. 

\subsection{GaussianAvatars}
Due to the capability of 3DGS to represent fine geometric structures, it has proved to be an efficient representation for creating photorealistic head avatars, as shown by GaussianAvatars~\cite{qian2023gaussianavatars}. GaussianAvatars proposes a method for dynamic 3D representation of human heads based on 3DGS by rigging the anisotropic 3D Gaussians to the faces of a 3D morphable face model. Specifically, the method uses FLAME~\cite{flame_siggraphAsia2017} as 3DMM due to its flexibility and compactness, consisting of only 5023 vertices and 9976 faces. To better represent mouth interior, it generated additional 120 vertices for teeth. Given a FLAME mesh, the idea is to first initialize a 3D Gaussian at the center of each triangle of the FLAME mesh and let the 3D Gaussian move with the faces of the FLAME mesh across different timesteps. 
For the paired 3D Gaussians with the faces of the FLAME mesh, the position $\boldsymbol{\mu}$, rotation $\boldsymbol{r}$ and anisotropic scaling $\boldsymbol{s}$ are defined in local space. During rendering, these are converted to global space as: 
\begin{equation}
    \boldsymbol{r'} = \boldsymbol{Rr},
\end{equation}
\begin{equation}
    \boldsymbol{\mu'} = k \boldsymbol{R\mu} + \boldsymbol{T}
\end{equation}
\begin{equation}
    \boldsymbol{s'} = k\boldsymbol{s},
\end{equation}
where $\boldsymbol{T}$ refers to the mean positions of the vertices of the triangle mesh, rotation matrix $\boldsymbol{R}$ describes the orientation of the triangles in the global space, scalar $k$ describes the triangle scaling. During avatar optimization, similar to 3DGS, the method uses adaptive density control strategy to add and remove splats based view-space positional gradient and opacity of each Gaussian. To prevent excessive pruning, the method also ensures that every triangle has at least one splat attached. The 3DGS parameters are then optimized using photometric loss $\mathcal{L}_\textrm{rgb}$, position loss $\mathcal{L}_\textrm{position}$ and scaling loss as $\mathcal{L}_\textrm{scale}$ as:
\begin{equation}
    \mathcal{L} = \mathcal{L}_\textrm{rgb} + \lambda_\textrm{pos} \mathcal{L}_\textrm{position} +  \lambda_\textrm{s} \mathcal{L}_\textrm{scaling}
\end{equation}
where $\mathcal{L}_\textrm{rgb}$ is combination of $\mathcal{L}_{1}$ and D-SSIM loss~\cite{kerbl3Dgaussians} between rendered and ground truth images.
\begin{equation}
    \mathcal{L}_\textrm{rgb} = (1-\lambda)\mathcal{L}_\textrm{1} + \lambda \mathcal{L}_\textrm{D-SSIM},
    \label{eq:rgb_loss}
\end{equation}
$\mathcal{L}_\textrm{position}$ ensures that splats remain close to their parent triangles: \begin{equation}
    \mathcal{L}_\textrm{position} = \big|\big| \max(\boldsymbol{\mu}, \boldsymbol{\epsilon}_\textrm{position})  \big|\big|_2,
\end{equation}
and $\mathcal{L}_\textrm{scaling}$ prevents excessive scaling of the splats: \begin{equation}
    \mathcal{L}_\textrm{scaling} = \big|\big| \max(\boldsymbol{s}, \boldsymbol{\epsilon}_\textrm{scaling})  \big|\big|_2,
\end{equation}

\section{Baselines}\label{sec:baselines}
We compare our method against audio-conditioned NeRF, 3DGS and mesh based methods. Since mesh-based methods can't generate photorealistic avatars, we combined audio-to-mesh methods with recent state-of-the-art mesh-to-3D avatar method GaussianAvatars~\cite{qian2023gaussianavatars}. Current NeRF and 3DGS based methods are designed for monocular videos only, thus in the main paper we train these methods only on the front cameras recording from our dataset. We breifly describe these methods as follows:

\textbf{Faceformer~\cite{faceformer2022}.} Faceformer leverages Wav2Vec2.0 to encode audio features, which are then processed by transformer-based autoregressive model via cross-modal multi-head attention to synthesize mesh animations. The method additionally uses biased causal multi-head self attention and  periodic positional encoding to improve generalization to longer sequences. The paper proposes a generic sequence model for a fixed set of identities with a style embedding to learn identity-specific speaking style.

\textbf{Codetalker~\cite{xing2023codetalker}.} Given an audio signal, Codetalker formulates speech-driven facial animation as code query task of a learnt codebook. The codebook is learnt by self-reconstruction of the mesh sequences with VQ-VAE. The learnt discrete codebook is then leveraged by code-query based temporal autoregressive model for speech-conditioned facial animation. The discrete motion space of finite cardinality can accurately audio-conditioned mesh animation. Similar to Faceformer, this method also encodes audio with Wav2Vec2.0.

\textbf{Imitator~\cite{Thambiraja_2023_ICCV}.} Similar to ours, Imitator learns a personalized model for speech-conditioned 3D facial animation. The method first pretrains a transformer based sequence model on high-quality VOCA dataset~\cite{VOCA2019} and then finetunes it with short Flame tracked sequences of the personalized avatar. To model lip closures accurately, the paper further proposes a novel lip contact loss based on physiological cues of bilabial consonants `m', `b' `p').

\textbf{RAD-NeRF~\cite{tang2022radnerf}.} The paper proposes audio-conditioned neural radiance fields for real-time rendering. The key contribution of the method is an efficient NeRF architecture by decomposing the video representation into three low-dimensional trainable feature grids. The first two feature grid model the audio and dynamic head motion respectively. The third feature grid models the torso motion. Compared to previous audio-conditioned NeRFs, this runs much faster and enables real-time inference. 

\textbf{ER-NeRF~\cite{li2023ernerf}.} Similar to RAD-NeRF, this paper also focusses on real-time rendering for audio-conditioned neural fields. However in contrast to RAD-NeRF, this method leverages triplane hash representation for learning spatial features. The method further captures the impact of audio features on different facial regions via region-aware attention module. To handle eye blinks, it uses explicit eye blinking control with a scalar. To model the torso, the method transforms a set of trainable 3D keypoints to normalized 2D coordinates and queries 2D neural field to predict the torso image.

\textbf{SyncTalk~\cite{peng2023synctalk}.} Building upon ER-NeRF, this method uses triplane hash representation and further improves the quality of lip synchronization. The method leverages face-synchronization controller that align lip motion with the corresponding audio signal and uses 3D facial blendshapes for capturing facial expressions. To model head pose, it utilizes 3D head tracker and stabilizes the head pose with a pretrained optical flow estimation model. Finally, it uses a portrait-sync generator to restore rest of the details like hair and background.

\textbf{GaussianTalker~\cite{GaussianTalker2024}.} The paper proposes audio-conditioned talking head generation framework based on 3D Gaussian Splatting (3DGS). By leveraging the speed and efficiency of 3DGS, the authors construct a canonical 3DGS representation of the head and deform it in synchronization with the audio during audio-driven animation. Specifically, it encodes spatial information (3DGS position) of the head via multiresolution triplane feature grid and uses an MLP for predicting rest of the 3DGS attributes in canonical space. Finally, these Gaussian attributes are merged with audio features via the Spatial-Audio attention that predict per-frame 3DGS deformations, enabling stability and control.

\textbf{TalkingGaussian~\cite{talkingGaussian2024}.} It is a deformation-based talking head synthesis framework, also leveraging 3DGS addressing the problem of facial distortion in existing radiance field methods. The authors represent dynamic talking head with deformable 3D Gaussians, and consists of (a) static persistent Gaussians representing persistent head structure and (b) neural grid-based motion field to handle dynamic facial motion. The model is decomposed into two branches to handle face and mouth interior separately with the goal to reconstruct more accurate motion and structure of mouth region.

\section{Dataset Details}\label{sec:dataset_details}
\begin{figure}[h!]
     \centering
     \includegraphics[width=1.0\linewidth]{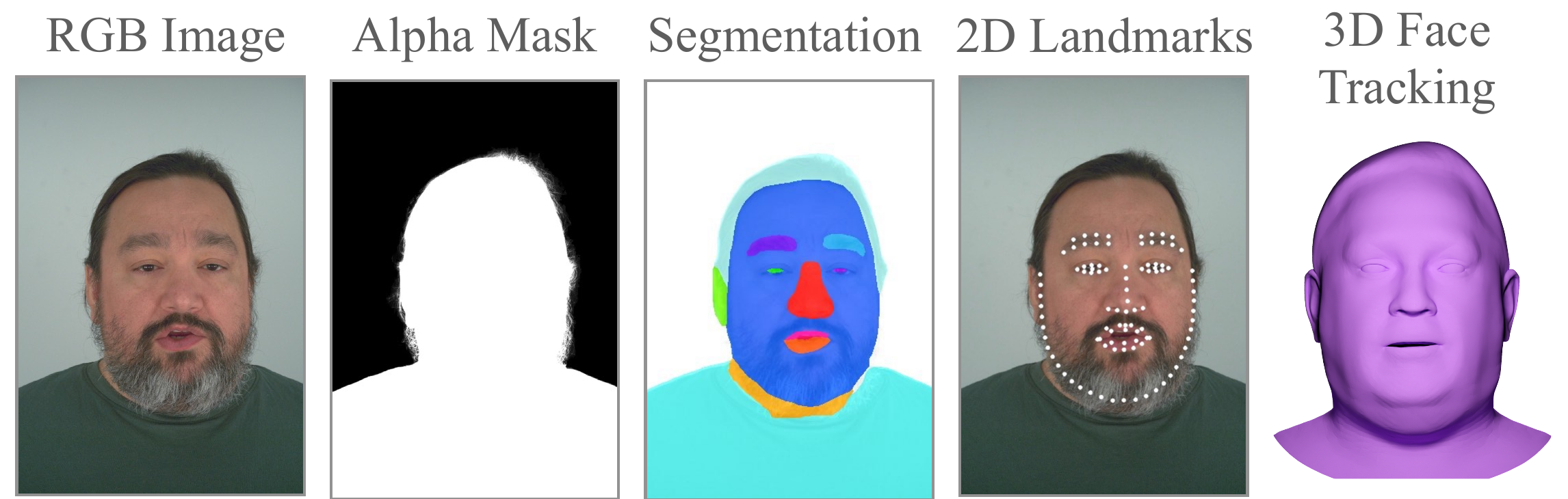}
     \caption{Dataset Metadata: Left to right, RGB Image, Alpha mask, Face segmentation, 2D landmarks and FLAME-based 3D Face Tracking. We will provide all these with our dataset release.}
     \label{fig:dataset_metadata}
\end{figure}

\begin{figure}[h!]
     \centering
     \includegraphics[width=0.9\linewidth]{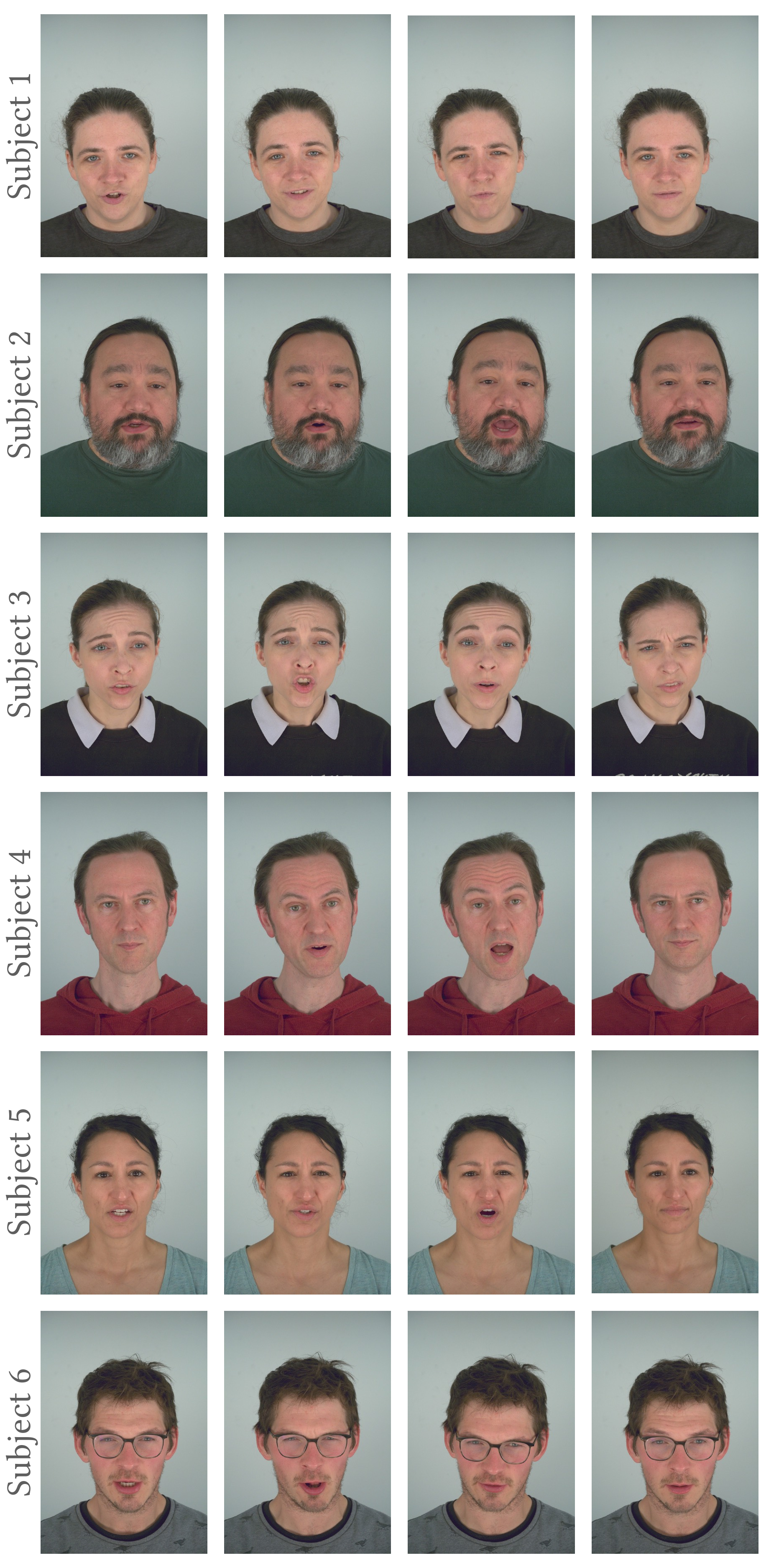}
     \caption{Dataset Participants: Four randomly selected frames from the dataset for each of six participants. The participants spoke with facial expressions and movements based on sentence transcript.}
     \label{fig:dataset_grid}
 \end{figure}

Our multi-view dataset consists of native speakers in age group 20-50 and includes three male and female participants, see Tab.~\ref{tab:participant_details}. We show additional metadata to be released with our dataset in Fig.~\ref{fig:dataset_metadata} and  some example frames from one of the sequences from our dataset in Fig.~\ref{fig:dataset_grid}. 

\begin{table}[h!]

     \begin{center}
     \begin{tabular}{c c c c}
         \toprule
         Participant & {Native Accent} &  {Age} & {Gender}  \\
         \toprule
         Subject 1 & British & 27 & Female \\
         Subject 2 & American & 42 & Male \\
         Subject 3 & American & 32  & Female \\
         Subject 4 & British  & 48  & Male \\
         Subject 5 & Canadian & 41  & Female \\
         Subject 6 & American  & 32  & Male  \\        
     \bottomrule
     \end{tabular}
     \end{center}
      \caption{Participant Details. We captured a gender-balanced dataset of six native English participants over a wide range of accents and age groups.}
     \label{tab:participant_details}
 \end{table}

 \begin{figure*}[t!]
     \centering
     \includegraphics[width=1.0\linewidth]{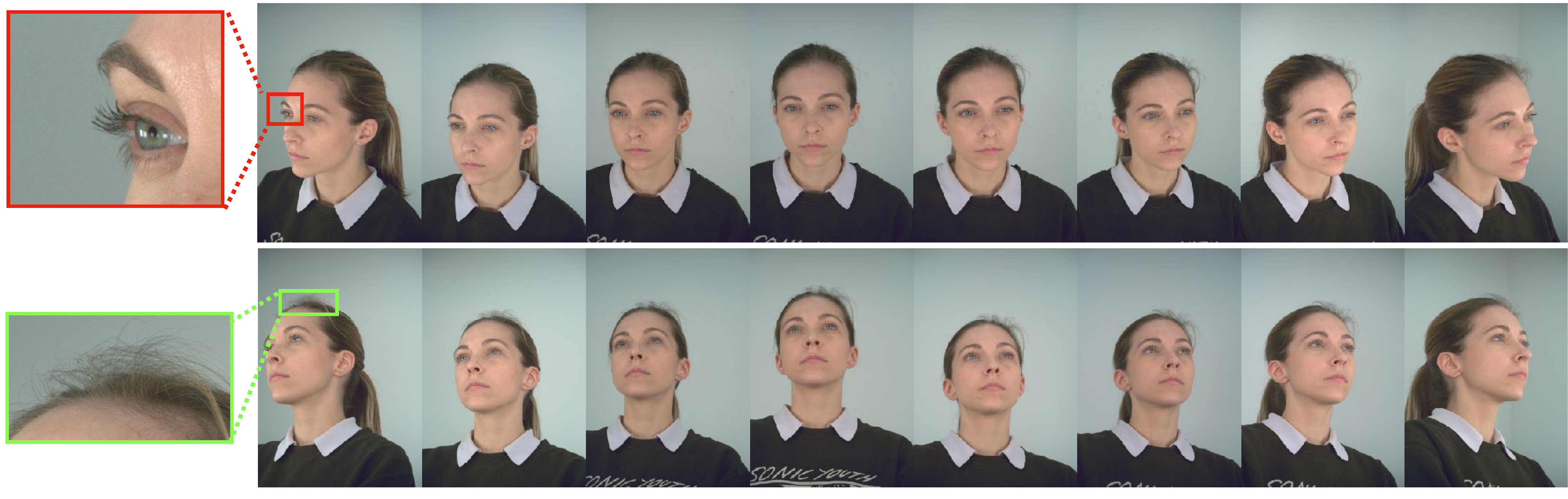}
     \caption{The multiview setup with 16 cameras used for recording participants captured at a resolution of 7.1 megapixels covering a field of view of 90$^{\circ}$ left-to-right and 30$^{\circ}$ up-to-down of the participant (right). The zoom-ins (left) for the eyes and hair show the level of detail captured by
     the cameras. Our audio-visual dataset contains detailed facial geometry, 
     }
\label{fig:dataset_multiview}
 \end{figure*}

\subsection{Hardware Configuration} We employ 16 machine vision cameras at a resolution of 7.1 megapixels and a supercardioid microphone to capture high-quality audio. Our capture setup is similar to Kirschstein \etal~\cite{kirschstein2023nersemble} covering a field of view of 90$^{\circ}$ left-to-right and 30$^{\circ}$ up-to-down. To avoid motion blur, we set the cameras at a shutter speed of 3ms. To capture the participant in appropriate lighting, we illuminate the subject with 8 LED light panels and use diffuser plates to reduce specularities on the skin.  Our dataset can capture fine-scale facial details like eyelashes, see Fig.~\ref{fig:dataset_multiview} for zoom-ins.

 \subsection{Speech Corpus} To maximize phonetic diversity in the dataset and to advance the field of audio-driven facial animation, we asked the participants to speak a phonetically diverse set of spoken English sentences carefully chosen from TIMIT speech corpus~\cite{timit} with expressions. We record the following categories of sentences from TIMIT corpus (a) Two accent-specific sentences that differ for people with different dialects, (b) 260 phonetically compact sentences to include a diverse range of phonetic contexts (c) 143 phonetically balanced sentences to include a balanced representation of phonemes. These short sentences range from 3-7 seconds each. Finally, we also record 10 free-form long sentences, where we ask participants a fixed set of questions based on their hobby/profession, etc, to capture the free form speaking style of the participant. These sentences are 10-20 seconds long. 
 \vspace{0.5cm}

For long sequences, the participants were asked 10 basic questions as listed below. 

\begin{enumerate}
    \item \texttt{Talk a little bit about your profession or education.}
    \item \texttt{Tell us about a recent trip or vacation you took.}
    \item \texttt{Share a hobby or activity that you enjoy pursuing in your free time.}
    \item \texttt{Describe a cultural event or festival you attended and what made it memorable.}
    \item \texttt{Describe a cuisine or dish you recently tried for the first time and your thoughts on it.}
    \item \texttt{Discuss a recent technological advancement or innovation that caught your attention.}
    \item \texttt{Talk about favourite movie/TV show.}
    \item \texttt{Who’s your favourite actor/singer?}
    \item \texttt{What’s your favourite sport?}
    \item \texttt{Your favourite holiday destination?}
\end{enumerate}

\vspace{0.5cm}
For short sequences, we recorded the following sentences from TIMIT corpus.

\begin{enumerate}
\item \texttt{She had your dark suit in greasy wash water all year.}
\item \texttt{Don't ask me to carry an oily rag like that.}
\item \texttt{Jane may earn more money by working hard.}
\item \texttt{Bright sunshine shimmers on the ocean.}
\item \texttt{Nothing is as offensive as innocence.}
\item \texttt{Why yell or worry over silly items?}
\item \texttt{Are your grades higher or lower than Nancy's?}
\item \texttt{Swing your arm as high as you can.}
\item \texttt{Before Thursday's exam, review every formula.}
\item \texttt{The museum hires musicians every evening.}
\item \texttt{Alimony harms a divorced man's wealth.}
\item \texttt{Aluminum silverware can often be flimsy.}
\item \texttt{She wore warm, fleecy, woolen overalls.}
\item \texttt{Those musicians harmonize marvelously.}
\item \texttt{Most young rise early every morning.}
\item \texttt{Beg that guard for one gallon of gas.}
\item \texttt{Help Greg to pick a peck of potatoes.}
\item \texttt{It's fun to roast marshmallows on a gas burner.}
\item \texttt{Coconut cream pie makes a nice dessert.}
\item \texttt{Only the most accomplished artists obtain popularity.}
\item \texttt{Critical equipment needs proper maintenance.}
\item \texttt{Young people participate in athletic activities.}
\item \texttt{Barb's gold bracelet was a graduation present.}
\item \texttt{Stimulating discussions keep students' attention.}
\item \texttt{Etiquette mandates compliance with existing regulations.}
\item \texttt{Biblical scholars argue history.}
\item \texttt{Addition and subtraction are learned skills.}
\item \texttt{That pickpocket was caught red-handed.}
\item \texttt{Grandmother outgrew her upbringing in petticoats.}
\item \texttt{At twilight on the twelfth day we'll have Chablis.}
\item \texttt{Catastrophic economic cutbacks neglect the poor.}
\item \texttt{Ambidextrous pickpockets accomplish more.}
\item \texttt{Her classical performance gained critical acclaim.}
\item \texttt{Even a simple vocabulary contains symbols.}
\item \texttt{The eastern coast is a place for pure pleasure and excitement.}
\item \texttt{The lack of heat compounded the tenant's grievances.}
\item \texttt{Academic aptitude guarantees your diploma.}
\item \texttt{The prowler wore a ski mask for disguise.}
\item \texttt{We experience distress and frustration obtaining our degrees.}
\item \texttt{The legislature met to judge the state of public education.}
\item \texttt{Chocolate and roses never fail as a romantic gift.}
\item \texttt{Any contributions will be greatly appreciated.}
\item \texttt{Continental drift is a geological theory.}
\item \texttt{We got drenched from the uninterrupted rain.}
\item \texttt{Last year's gas shortage caused steep price increases.}
\item \texttt{Upgrade your status to reflect your wealth.}
\item \texttt{Eat your raisins outdoors on the porch steps.}
\item \texttt{Porcupines resemble sea urchins.}
\item \texttt{Cliff's display was misplaced on the screen.}
\item \texttt{An official deadline cannot be postponed.}
\item \texttt{Fill that canteen with fresh spring water.}
\item \texttt{Gently place Jim's foam sculpture in the box.}
\item \texttt{Bagpipes and bongos are musical instruments.}
\item \texttt{Doctors prescribe drugs too freely.}
\item \texttt{Will you please describe the idiotic predicament.}
\item \texttt{It's impossible to deal with bureaucracy.}
\item \texttt{Good service should be rewarded by big tips.}
\item \texttt{My instructions desperately need updating.}
\item \texttt{Cooperation along with understanding alleviate dispute.}
\item \texttt{Primitive tribes have an upbeat attitude.}
\item \texttt{Flying standby can be practical if you want to save money.}
\item \texttt{The misprint provoked an immediate disclaimer.}
\item \texttt{A large household needs lots of appliances.}
\item \texttt{Youngsters love common candy as treats.}
\item \texttt{Iguanas and alligators are tropical reptiles.}
\item \texttt{Masquerade parties tax one's imagination.}
\item \texttt{Penguins live near the icy Antarctic.}
\item \texttt{Medieval society was based on hierarchies.}
\item \texttt{Project development was proceeding too slowly.}
\item \texttt{Kindergarten children decorate their classrooms for all holidays.}
\item \texttt{Special task forces rescue hostages from kidnappers.}
\item \texttt{Call an ambulance for medical assistance.}
\item \texttt{He stole a dime from a beggar.}
\item \texttt{A huge tapestry hung in her hallway.}
\item \texttt{Birthday parties have cupcakes and ice cream.}
\item \texttt{His scalp was blistered from today's hot sun.}
\item \texttt{She slipped and sprained her ankle on the steep slope.}
\item \texttt{The best way to learn is to solve extra problems.}
\item \texttt{Tugboats are capable of hauling huge loads.}
\item \texttt{A muscular abdomen is good for your back.}
\item \texttt{The cartoon features a muskrat and a tadpole.}
\item \texttt{The emblem depicts the Acropolis all aglow.}
\item \texttt{The mango and the papaya are in a bowl.}
\item \texttt{Combine all the ingredients in a large bowl.}
\item \texttt{The misquote was retracted with an apology.}
\item \texttt{The coyote, bobcat, and hyena are wild animals.}
\item \texttt{Trespassing is forbidden and subject to penalty.}
\item \texttt{Encyclopedias seldom present anecdotal evidence.}
\item \texttt{A screwdriver is made from vodka and orange juice.}
\item \texttt{Westchester is a county in New York.}
\item \texttt{Artificial intelligence is for real.}
\item \texttt{Lots of foreign movies have subtitles.}
\item \texttt{Angora cats are furrier than Siamese.}
\item \texttt{Publicity and notoriety go hand in hand.}
\item \texttt{Pizzerias are convenient for a quick lunch.}
\item \texttt{December and January are nice months to spend in Miami.}
\item \texttt{Technical writers can abbreviate in bibliographies.}
\item \texttt{Scientific progress comes from the development of new techniques.}
\item \texttt{Tradition requires parental approval for under-age marriage.}
\item \texttt{The clumsy customer spilled some expensive perfume.}
\item \texttt{The bungalow was pleasantly situated near the shore.}
\item \texttt{Pledge to participate in Nevada's aquatic competition.}
\item \texttt{Which long article was opaque and needed clarification?}
\item \texttt{The sound of Jennifer's bugle scared the antelope.}
\item \texttt{The willowy woman wore a muskrat coat.}
\item \texttt{Too much curiosity can get you into trouble.}
\item \texttt{Correct execution of my instructions is crucial.}
\item \texttt{Most precincts had a third of the votes counted.}
\item \texttt{While waiting for Chipper she crisscrossed the square many times.}
\item \texttt{The previous speaker presented ambiguous results.}
\item \texttt{Mosquitoes exist in warm, humid climates.}
\item \texttt{Scholastic aptitude is judged by standardized tests.}
\item \texttt{Orange juice tastes funny after toothpaste.}
\item \texttt{The water contained too much chlorine and stung his eyes.}
\item \texttt{Our experiment's positive outcome was unexpected.}
\item \texttt{Remove the splinter with a pair of tweezers.}
\item \texttt{The government sought authorization of his citizenship.}
\item \texttt{As coauthors, we presented our new book to the haughty audience.}
\item \texttt{As a precaution, the outlaws bought gunpowder for their stronghold.}
\item \texttt{Her auburn hair reminded him of autumn leaves.}
\item \texttt{They remained lifelong friends and companions.}
\item \texttt{Curiosity and mediocrity seldom coexist.}
\item \texttt{The easygoing zoologist relaxed throughout the voyage.}
\item \texttt{Biologists use radioactive isotopes to study microorganisms.}
\item \texttt{Employee layoffs coincided with the company's reorganization.}
\item \texttt{How would you evaluate this algebraic expression?}
\item \texttt{The Mayan neoclassic scholar disappeared while surveying ancient ruins.}
\item \texttt{The diagnosis was discouraging; however, he was not overly worried.}
\item \texttt{The triumphant warrior exhibited naive heroism.}
\item \texttt{Whoever cooperates in finding Nan's cameo will be rewarded.}
\item \texttt{The haunted house was a hit due to outstanding audio-visual effects.}
\item \texttt{Severe myopia contributed to Ron's inferiority complex.}
\item \texttt{Buying a thoroughbred horse requires intuition and expertise.}
\item \texttt{She encouraged her children to make their own Halloween costumes.}
\item \texttt{We could barely see the fjords through the snow flurries.}
\item \texttt{Almost all colleges are now coeducational.}
\item \texttt{Rich looked for spotted hyenas and jaguars on the safari.}
\item \texttt{Why else would Danny allow others to go?}
\item \texttt{Who authorized the unlimited expense account?}
\item \texttt{Destroy every file related to my audits.}
\item \texttt{Serve the coleslaw after I add the oil.}
\item \texttt{Withdraw all phony accusations at once.}
\item \texttt{Straw hats are out of fashion this year.}
\item \texttt{Draw each graph on a new axis.}
\item \texttt{Norwegian sweaters are made of lamb's wool.}
\item \texttt{Young children should avoid exposure to contagious diseases.}
\item \texttt{Ralph controlled the stopwatch from the bleachers.}
\item \texttt{Approach your interview with statuesque composure.}
\item \texttt{The causeway ended abruptly at the shore.}
\item \texttt{Even I occasionally get the Monday blues!}
\item \texttt{Military personnel are expected to obey government orders.}
\item \texttt{When peeling an orange, it is hard not to spray juice.}
\item \texttt{Rob sat by the pond and sketched the stray geese.}
\item \texttt{Michael colored the bedroom wall with crayons.}
\item \texttt{I gave them several choices and let them set the priorities.}
\item \texttt{The news agency hired a great journalist.}
\item \texttt{The morning dew on the spider web glistened in the sun.}
\item \texttt{The sermon emphasized the need for affirmative action.}
\item \texttt{The small boy put the worm on the hook.}
\item \texttt{Try to recall the events in chronological order.}
\item \texttt{Nonprofit organizations have frequent fund raisers.}
\item \texttt{The most recent geological survey found seismic activity.}
\item \texttt{Cory attacked the project with extra determination.}
\item \texttt{You always come up with pathological examples.}
\item \texttt{Put the butcher block table in the garage.}
\item \texttt{Keep the thermometer under your tongue!}
\item \texttt{Steph could barely handle the psychological trauma.}
\item \texttt{It's healthier to cook without sugar.}
\item \texttt{Allow leeway here, but rationalize all errors.}
\item \texttt{His failure to open the store by eight cost him his job.}
\item \texttt{Highway and freeway mean the same thing.}
\item \texttt{The paper boy bought two apples and three ices.}
\item \texttt{Clear pronunciation is appreciated.}
\item \texttt{A doctor was in the ambulance with the patient.}
\item \texttt{Puree some fruit before preparing the skewers.}
\item \texttt{It's not easy to create illuminating examples.}
\item \texttt{The hallway opens into a huge chamber.}
\item \texttt{May I order a strawberry sundae after I eat dinner?}
\item \texttt{They all agree that the essay is barely intelligible.}
\item \texttt{Herb's birthday occurs frequently on Thanksgiving.}
\item \texttt{The cigarettes in the clay ashtray overflowed onto the oak table.}
\item \texttt{Reading in poor light gives you eyestrain.}
\item \texttt{The Boston Ballet overcame their funding shortage.}
\item \texttt{We apply auditory modeling to computer speech recognition.}
\item \texttt{The gorgeous butterfly ate a lot of nectar.}
\item \texttt{Tornados often destroy acres of farm land.}
\item \texttt{Remember to allow identical twins to enter freely.}
\item \texttt{How oily do you like your salad dressing?}
\item \texttt{We saw eight tiny icicles below our roof.}
\item \texttt{The saw is broken, so chop the wood instead.}
\item \texttt{Withdraw only as much money as you need.}
\item \texttt{Draw every outer line first, then fill in the interior.}
\item \texttt{The jaw operates by using antagonistic muscles.}
\item \texttt{Cliff was soothed by the luxurious massage.}
\item \texttt{Steve wore a bright red cashmere sweater.}
\item \texttt{To further his prestige, he occasionally reads the Wall Street Journal.}
\item \texttt{Alice's ability to work without supervision is noteworthy.}
\item \texttt{Cory and Trish played tag with beach balls for hours.}
\item \texttt{The tooth fairy forgot to come when Roger's tooth fell out.}
\item \texttt{Planned parenthood organizations promote birth control.}
\item \texttt{Jeff thought you argued in favor of a centrifuge purchase.}
\item \texttt{Rich purchased several signed lithographs.}
\item \texttt{In every major cloverleaf, traffic sometimes gets backed up.}
\item \texttt{In the long run, it pays to buy quality clothing.}
\item \texttt{Brush fires are common in the dry underbrush of Nevada.}
\item \texttt{Weatherproof galoshes are very useful in Seattle.}
\item \texttt{This brochure is particularly informative for a prospective buyer.}
\item \texttt{The avalanche triggered a minor earthquake.}
\item \texttt{These exclusive documents must be locked up at all times.}
\item \texttt{Please take this dirty table cloth to the cleaners for me.}
\item \texttt{Should giraffes be kept in small zoos?}
\item \texttt{If Carol comes tomorrow, have her arrange for a meeting at two.}
\item \texttt{I'd rather not buy these shoes than be overcharged.}
\item \texttt{Shaving cream is a popular item on Halloween.}
\item \texttt{Amoebas change shape constantly.}
\item \texttt{We like bleu cheese but Victor prefers swiss cheese.}
\item \texttt{Tofu is made from processed soybeans.}
\item \texttt{The bluejay flew over the high building.}
\item \texttt{Cheap stockings run the first time they're worn.}
\item \texttt{Cottage cheese with chives is delicious.}
\item \texttt{Shipbuilding is a most fascinating process.}
\item \texttt{The proof that you are seeking is not available in books.}
\item \texttt{The hood of the jeep was steaming in the hot sun.}
\item \texttt{My desires are simple: give me one informative paragraph on the subject.}
\item \texttt{Those answers will be straightforward if you think them through carefully first.}
\item \texttt{If people were more generous, there would be no need for welfare.}
\item \texttt{The nearest synagogue may not be within walking distance.}
\item \texttt{The groundhog clearly saw his shadow, but stayed out only a moment.}
\item \texttt{The local drugstore was charged with illegally dispensing tranquilizers.}
\item \texttt{Al received a joint appointment in the biology and the engineering departments.}
\item \texttt{Gregory and Tom chose to watch cartoons in the afternoon.}
\item \texttt{Chip postponed alimony payments until the latest possible date.}
\item \texttt{Count the number of teaspoons of soysauce that you add.}
\item \texttt{The big dog loved to chew on the old rag doll.}
\item \texttt{Todd placed top priority on getting his bike fixed.}
\item \texttt{An adult male baboon's teeth are not suitable for eating shellfish.}
\item \texttt{Often you'll get back more than you put in.}
\item \texttt{Gus saw pine trees and redwoods on his walk through Sequoia National Forest.}
\item \texttt{Rob made Hungarian goulash for dinner and gooseberry pie for dessert.}
\item \texttt{Bob bandaged both wounds with the skill of a doctor.}
\item \texttt{The high security prison was surrounded by barbed wire.}
\item \texttt{Take charge of choosing her bride's maids' gowns.}
\item \texttt{The frightened child was gently subdued by his big brother.}
\item \texttt{The barracuda recoiled from the serpent's poisonous fangs.}
\item \texttt{The patient and the surgeon are both recuperating from the lengthy operation.}
\item \texttt{I'll have a scoop of that exotic purple and turquoise sherbet.}
\item \texttt{The preschooler couldn't verbalize her feelings about the emergency conditions.}
\item \texttt{Many wealthy tycoons splurged and bought both a yacht and a schooner.}
\item \texttt{The new suburbanites worked hard on refurbishing their older home.}
\item \texttt{According to my interpretation of the problem, two lines must be perpendicular.}
\item \texttt{The system may break down soon, so save your files frequently.}
\item \texttt{The annoying raccoons slipped into Phil's garden every night.}
\item \texttt{I took her word for it, but is she really going with you?}
\item \texttt{The gunman kept his victim cornered at gunpoint for three hours.}
\item \texttt{Will you please confirm government policy regarding waste removal?}
\item \texttt{The fish began to leap frantically on the surface of the small lake.}
\item \texttt{Her wardrobe consists of only skirts and blouses.}
\item \texttt{There was a gigantic wasp next to Irving's big top hat.}
\item \texttt{Those who are not purists use canned vegetables when making stew.}
\item \texttt{They used an aggressive policeman to flag thoughtless motorists.}
\item \texttt{Shell shock caused by shrapnel is sometimes cured through group therapy.}
\item \texttt{Ralph prepared red snapper with fresh lemon sauce for dinner.}
\item \texttt{If you destroy confidence in banks, you do something to the economy, he said.}
\item \texttt{He further proposed grants of an unspecified sum for experimental hospitals.}
\item \texttt{Nothing has been done yet to take advantage of the enabling legislation.}
\item \texttt{It also provides for funds to clear slums and help colleges build dormitories.}
\item \texttt{The prospect of cutting back spending is an unpleasant one for any governor.}
\item \texttt{He really crucified him; he nailed it for a yard loss.}
\item \texttt{There is definitely some ligament damage in his knee.}
\item \texttt{In fact our whole defensive unit did a good job.}
\item \texttt{He played basketball there while working toward a law degree.}
\item \texttt{So, if anybody solicits by phone, make sure you mail the dough to the above.}
\item \texttt{Her position covers a number of daily tasks common to any social director.}
\item \texttt{The structures housing the apartments are of masonry and frame construction.}
\item \texttt{This, he added, brought about petty jealousies and petty personal grievances.}
\item \texttt{There was no confirmation of such massive assaults from independent sources.}
\item \texttt{The staff deserves a lot of credit working down here under real obstacles.}
\item \texttt{They make gin saws and deal in parts, supplies and some used gin machinery.}
\item \texttt{Maybe it's taking longer to get things squared away than the bankers expected.}
\item \texttt{Hiring the wife for one's company may win her tax-aided retirement income.}
\item \texttt{Unfortunately, there is still little demand for broccoli and cauliflower.}
\item \texttt{Displayed as lamps, the puppets delight the children and are decorative accent.}
\item \texttt{To create such a lamp, order a wired pedestal from any lamp shop.}
\item \texttt{There are more obvious nymphomaniacs on any private-eye series.}
\item \texttt{But this doesn't detract from its merit as an interesting, if not great, film.}
\item \texttt{And you think you have language problems.}
\item \texttt{Ideally, he knew, it should be preceded by concrete progress at lower levels.}
\item \texttt{This is a significant advance but its import should not be exaggerated.}
\item \texttt{Adequate compensation is indispensable.}
\item \texttt{This is a problem that goes considerably beyond questions of salary and tenure.}
\item \texttt{Some observers speculated that this might be his revenge on his home town.}
\item \texttt{Confusion became chaos; each succeeding day brought new acts of violence.}
\item \texttt{That added traffic means rising streams of dimes and quarters at toll gates.}
\item \texttt{Traffic frequently has failed to measure up to engineers' rosy estimates.}
\item \texttt{Progress is being made, too, in improving motorists' access to many turnpikes.}
\item \texttt{Under this law annual grants are given to systems in substantial amounts.}
\item \texttt{Within a system, however, the autonomy of each member library is preserved.}
\item \texttt{The desire and ability to read are important aspects of our cultural life.}
\item \texttt{We congratulate the entire membership on its record of good legislation.}
\item \texttt{Thereupon followed a demonstration that tyranny knows no ideological confines.}
\item \texttt{Wooded stream valleys in the folds of earth would be saved.}
\item \texttt{His election, on the other hand, would unquestionably strengthen the regulars.}
\item \texttt{He spoke briefly, sensibly, to the point and without oratorical flourishes.}
\item \texttt{Further, it has its work cut out stopping anarchy where it is now garrisoned.}
\item \texttt{Fools, he bayed, what do you think you are doing?}
\item \texttt{So we note approvingly a fresh sample of unanimity.}
\item \texttt{It is one of the rare public ventures here on which nearly everyone is agreed.}
\item \texttt{Thus there is a clearer division of authority, administrative and legislative.}
\item \texttt{Jokes, cartoons and cynics to the contrary, mothers-in-law make good friends.}
\item \texttt{Theirs is a sacrificial life by earthly standards.}
\item \texttt{The narrow fringe of sadness that ran around it only emphasized the pleasure.}
\item \texttt{Would a blue feather in a man's hat make him happy all day?}
\item \texttt{These programs emphasize the acceptance of biracial classrooms peacefully.}
\item \texttt{You certainly can't expect the infield to do any better than it did last year.}
\item \texttt{Is the mother of an autistic child at fault?}
\item \texttt{As a rule, the autistic child doesn't enjoy physical contact with others.}
\item \texttt{Or certain words or rituals that child and adult go through may do the trick.}
\item \texttt{We did not accept the diagnosis at once, but gradually we are coming to.}
\item \texttt{Is a relaxed home atmosphere enough to help her outgrow these traits?}
\item \texttt{Where only one club existed before, he says, two will flourish henceforth.}
\item \texttt{This is going to be a language lesson, and you can master it in a few minutes.}
\item \texttt{Family loyalties and cooperative work have been unbroken for generations.}
\item \texttt{Heels place emphasis on the long legged silhouette.}
\item \texttt{Wine glass heels are to be found in both high and semi-heights.}
\item \texttt{Stacked heels are also popular on dressy or tailored shoes.}
\item \texttt{Contrast trim provides other touches of color.}
\item \texttt{At the left is a pair of dressy straw pumps in a light, but crisp texture.}
\item \texttt{At right is a casual style in a crushed unlined white leather.}
\item \texttt{Most of us brush our teeth by hand.}
\item \texttt{The bristles are soft enough to massage the gums and not scratch the enamel.}
\item \texttt{"Steam baths" writes: do steam baths have any health value?}
\item \texttt{"Sewing brings numbness" writes: what makes my hands numb when sewing?}
\item \texttt{Teaching guides are included with each record.}
\item \texttt{He doesn't want her to look frowningly at him, or speak to him angrily.}
\item \texttt{But even mother's loving attitude will not always prevent misbehavior.}
\item \texttt{She can decrease the number of temptations.}
\item \texttt{She can remove all knick-knacks within reach.}
\item \texttt{Usually, they titter loudly after they have passed by.}
\item \texttt{Too often, unless he hails them, they pass him by.}
\item \texttt{Say he is a horse thief, runs an old adage.}
\item \texttt{It seems that open season upon veterans' hospitalization is once more upon us.}
\item \texttt{This we can sympathetically understand.}
\item \texttt{This is taxation without representation.}
\item \texttt{Our entire economy will have a terrific uplift.}
\item \texttt{One even gave my little dog a biscuit.}
\item \texttt{Maybe he will help to turn our fair city into a ghost town.}
\item \texttt{Why do we need bigger and better bombs?}
\item \texttt{One of the problems associated with the expressway stems from the basic idea.}
\item \texttt{Bridges, tunnels and ferries are the most common methods of river crossings.}
\item \texttt{Replace it with the statue of one or another of the world's famous dictators.}
\item \texttt{The gallant half-city is dying on its feet.}
\item \texttt{Their privations are almost beyond endurance.}
\item \texttt{The moment of truth is the moment of crisis.}
\item \texttt{New self-deceiving rags are hurriedly tossed on the too-naked bones.}
\item \texttt{What explains this uni-directional paralysis?}
\item \texttt{Originals are not necessarily good and adaptations are not necessarily bad.}
\item \texttt{But that explanation is only partly true.}
\item \texttt{But the ships are very slow now, and we don't get so many sailors any more.}
\item \texttt{They were shown how to advance against an enemy outpost atop a cleared ridge.}
\item \texttt{We would lose our export markets and deny ourselves the imports we need.}
\item \texttt{With this no loyal citizen can quarrel.}
\item \texttt{This possibility is anything but reassuring.}
\item \texttt{The public is now armed with sophistication and numerous competing media.}
\item \texttt{But the attack was made from an advance copy.}
\item \texttt{Well, now we have two big theaters.}
\item \texttt{Splendor by sorcery: it's a horror.}
\item \texttt{One-upmanship is practiced by both sides in a total war.}
\item \texttt{This big, flexible voice with uncommon range has been superbly disciplined.}
\item \texttt{Her debut over, perhaps the earlier scenes will emerge equally fine.}
\item \texttt{He injected more vitality into the score than it has revealed in many years.}
\item \texttt{The storyline, in sort, is wildly unrealistic.}
\item \texttt{He talked about unauthentic storylines too.}
\item \texttt{He praises many individuals generously.}
\item \texttt{His portrayal of an edgy head-in-the-clouds artist is virtually flawless.}
\item \texttt{Not a corner has been visibly cut in this one.}
\item \texttt{The master's hand has lost none of its craft.}
\item \texttt{He showed puny men attacked by splendidly tyrannical machines.}
\item \texttt{He may have a point in urging that decadent themes be given fewer prizes.}
\item \texttt{The works are presented chronologically.}
\item \texttt{The humor of the situation can be imagined.}
\item \texttt{His technique is ample and his musical ideas are projected beautifully.}
\item \texttt{What a discussion can ensue when the title of this type of song is in question.}
\item \texttt{Program note reads as follows: take hands; this urgent visage beckons us.}
\item \texttt{The orchestra was obviously on its mettle and it played most responsively.}
\item \texttt{He liked to nip ear lobes of unsuspecting visitors with his needle-sharp teeth.}
\item \texttt{Here, he is, quite persuasively, the very embodiment of meanness and slyness.}
\item \texttt{He is a man of major talent -- but a man of solitary, uncertain impulses.}
\item \texttt{He was above all a friend seeker, almost pathetic in his eagerness to be liked.}
\item \texttt{He enlisted a staff of loyal experts and of many zealous volunteers.}
\item \texttt{But what has been happening recently might be described as creeping mannerism.}
\item \texttt{Clever light songs were overly coy, tragic songs a little too melodramatic.}
\item \texttt{Below is a specific guide, keyed to the calendar.}
\item \texttt{A sailboat may have a bone in her teeth one minute and lie becalmed the next.}
\item \texttt{It suffers from a lack of unity of purpose and respect for heroic leadership.}
\item \texttt{The fat man has trouble buying life insurance or has to pay higher premiums.}
\item \texttt{Far more frequently, overeating is the result of a psychological compulsion.}
\item \texttt{Yet it exists and has an objective reality which can be experienced and known.}
\item \texttt{Yet the spirit which lives in community is not identical with the community.}
\item \texttt{But this statement is completely unconvincing.}
\item \texttt{A second point requires more extended comment.}
\item \texttt{The straight line would symbolize its uniqueness, the circle its universality.}
\item \texttt{His history is his alone, yet each man must recognize his own history in it.}
\item \texttt{Death reminds man of his sin, but it reminds him also of his transience.}
\item \texttt{Such a calm and assuring peace can be yours.}
\item \texttt{Satellites, sputniks, rockets, balloons; what next?}

\end{enumerate}

\clearpage 

\end{document}